\documentclass{article} 
\usepackage{iclr2023_conference,times}

\usepackage{amsmath,amsfonts,bm}

\def\eqref#1{equation~\ref{#1}}

\def\1{\bm{1}}

\DeclareMathAlphabet{\mathsfit}{\encodingdefault}{\sfdefault}{m}{sl}
\SetMathAlphabet{\mathsfit}{bold}{\encodingdefault}{\sfdefault}{bx}{n}

\usepackage{array}
\usepackage{hyperref}
\usepackage{url}
\usepackage{microtype}
\usepackage{graphicx}
\usepackage{subfigure}
\usepackage{amssymb}
\usepackage{amsfonts}
\usepackage{fdsymbol}
\usepackage{booktabs} 
\usepackage{multirow}
\usepackage{amsmath}
\usepackage{threeparttable}
\usepackage{appendix}
\newtheorem{theorem}{Theorem}
\newtheorem{definition}{Definition}
\usepackage{hyperref}

\usepackage{algorithm}
\usepackage{algorithmic}
\title{Generalizing Multimodal Variational Methods to Sets}
\author{Charles Zhou \& Yiqun Duan  \\
FEIT \\
University of Technology Sydney\\
Sydney, Australia \\
\texttt{\{Jinzhao.Zhou,Yiqun.Duan\}@stdent.uts.edu.au} \\
\And
Zhihong Chen\\
Department of Mathematics\\
South China University of Technology\\
Guangzhou, China \\
\texttt{mathematics.hong@gmail.com} \\
\AND
Yu-Cheng Chang, Chin-Teng Lin\\
University of Technology Sydney\\
Sydney, Australia \\
\texttt{yu-cheng.chang,chin-teng.lin\}@uts.edu.au}
}
\iclrfinalcopy 
\begin{document}
\maketitle
\begin{abstract}
Making sense of multiple modalities can yield a more comprehensive description of real-world phenomena. However, learning the co-representation of diverse modalities is still a long-standing endeavor in emerging machine learning applications and research. Previous generative approaches for multimodal input approximate a joint-modality posterior by uni-modality posteriors as product-of-experts (PoE) or mixture-of-experts (MoE). We argue that these approximations lead to a defective bound for the optimization process and loss of semantic connection among modalities. This paper presents a novel variational method on sets called the Set Multimodal VAE (SMVAE) for learning a multimodal latent space while handling the missing modality problem. By modeling the joint-modality posterior distribution directly, the proposed SMVAE learns to exchange information between multiple modalities and compensate for the drawbacks caused by factorization. In public datasets of various domains, the experimental results demonstrate that the proposed method is applicable to order-agnostic cross-modal generation while achieving outstanding performance compared to the state-of-the-art multimodal methods. The source code for our method is available online https://anonymous.4open.science/r/SMVAE-9B3C/.
\end{abstract}

\section{Introduction}\label{Introduction}
Most real-life applications such as robotic systems, social media mining, and recommendation systems naturally contain multiple data sources, which raise the need for learning co-representation among diverse modalities \cite{lee2020making}. Making use of additional modalities should improve the general performance of downstream tasks as it can provide more information from another perspective. In literatures, substantial improvements can be achieved by utilizing another modality as supplementary information \cite{asano2020labelling,nagrani2020speech2action} or by multimodal fusion \cite{atrey2010multimodal, hori2017attention, zhang2021deep}. However, current multimodal research suffers severely from the lack of multimodal data with fine-grained labeling and alignment \cite{sun2017revisiting,beyer2020we,rahate2022multimodal,baltruvsaitis2018challenges} and the missing of modalities \cite{ma2021smil, chen2021multi}. 

In the self-supervised and weakly-supervised learning field, the variational autoencoders (VAEs) for multimodal data \cite{kingma2013auto,wu2018multimodal,shi2019variational,sutter2021generalized} have been a dominating branch of development. VAEs are generative self-supervised models by definition that capture the dependency between an unobserved latent variable and the input observation. To jointly infer the latent representation and reconstruct the observations properly, the multimodal VAEs are required to extract both modality-specific and modality-invariant features from the multimodal observations. Earlier works mainly suffer from scalability issues as they need to learn a separate model for each modal combination \cite{pandey2017variational,yan2016attribute2image}. More recent multimodal VAEs handle this issue and achieves scalability by approximating the true joint posterior distribution with the mixture or the product of uni-modality inference models \cite{shi2019variational,wu2018multimodal,sutter2021generalized}. However, our key insight is that their methods suffer from two critical drawbacks: 1) The implied conditional independence assumption and corresponding factorization deviate their VAEs from modeling inter-modality correlations. 2) The aggregation of inference results from uni-modality is by no means a co-representation of these modalities.

\begin{figure*}[htbp]
        \centering
        \includegraphics[width=0.8\linewidth]{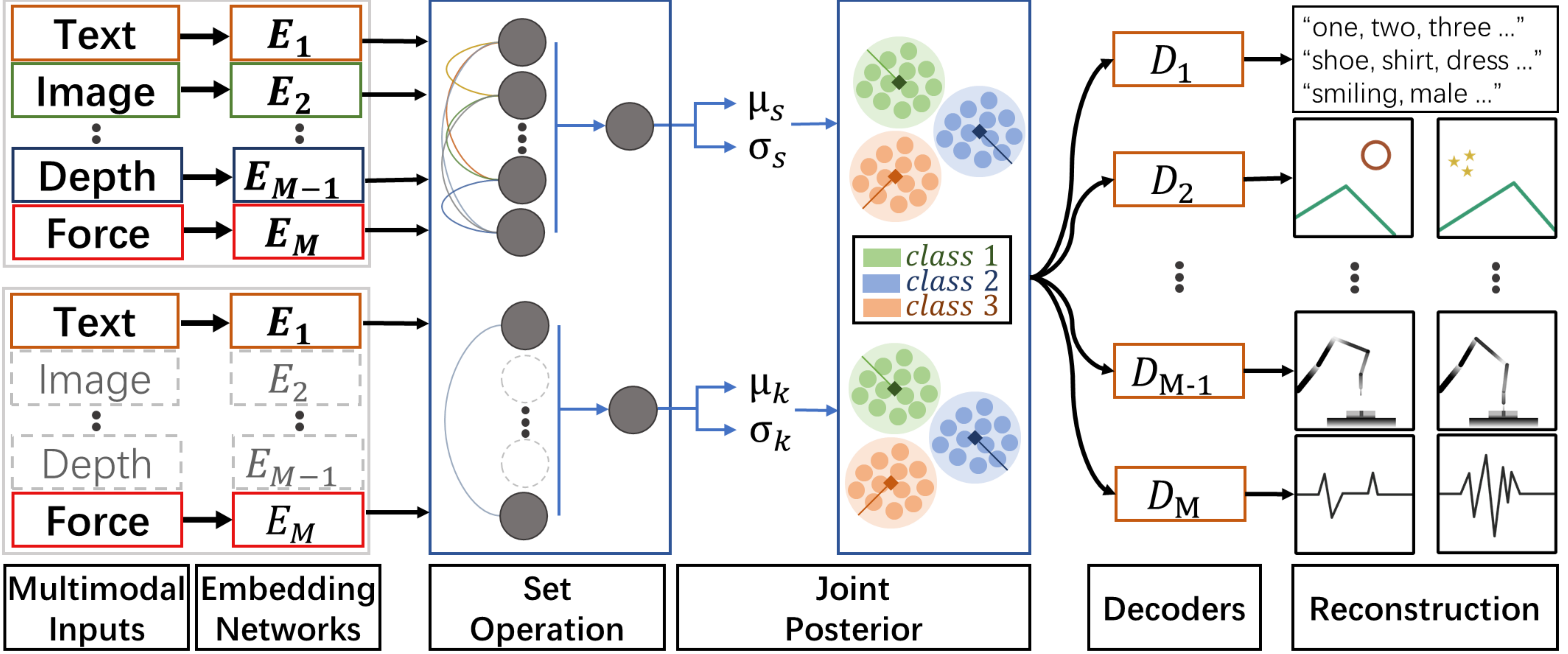}
        \caption{Overview of the proposed method for learning multimodal latent space. The SMVAE is able to handle any combination or number of input modalities while having discriminative latent space and proper reconstruction. }
        \label{fig:cover}
\end{figure*}

To overcome these drawbacks of previous VAE methods, this work proposes the Set Multimodal Variational Autoencoder (SMVAE), a novel multimodel generative model eschewing factorization and instead relying solely upon set operation to achieve scalability. The SMVAE allows for better performance compared to the latest multimodal VAE methods and can handle input modalities of variable numbers and permutations. By learning the actual multimodal joint posterior directly, the SMVAE is the first multimodal VAE method that achieves scalable co-representation with missing modalities. A high-level overview of the proposed method is illustrated in Fig.\ref{fig:cover}. The SMVAE can handle a set of maximally $M$ modalities as well as their subsets and allows cross-modality generations. $E_i$ and $D_i$ represent the $i-$th embedding network and decoder network for the specific modality. $\mu_s,\sigma_s$ and $\mu_k,\sigma_k$ represent the parameters for the posterior distribution of the latent variable. By incorporating set operation when learning the joint-modality posterior, we can simply drop the corresponding embedding networks when a modality is missing. Comprehensive experiments show the proposed Set Multimodal Variational Autoencoder (SMVAE) outperforms state-of-the-art multimodal VAE methods and is immediately applicable to real-life multimodality.
\section{Related Work}\label{Related}
\subsection{Multimodality VAEs}
The core problem of learning a multimodal generative model is to maintain the model's scalability to the exponential number of modal combinations. Existing multimodal generative models such as Conditional VAE (CVAE)\cite{pandey2017variational} and joint-modality VAE (JMVAE) \cite{suzuki2016joint} had difficulty scaling since they need to assign a separate inference model for each possible input and output combinations. To tackle this issue, follow-up works, such as, TELBO \cite{vedantam2017generative}, MVAE \cite{wu2018multimodal}, MMVAE \cite{shi2019variational}, MoPoE \cite{sutter2021generalized}, assume the variational approximation is factorizable. Thus, they focused on factorizing the approximation of the multimodal joint posterior $q(\boldsymbol{\mathrm{z}}|\boldsymbol{\mathrm{x}}_\mathrm{1}, \cdots, \boldsymbol{\mathrm{x}}_\mathrm{M})$ into a set of uni-modality inference encoders ${q_i(\boldsymbol{\mathrm{z}}|\boldsymbol{\mathrm{x}}_\mathrm{i})}$, such that $q(\boldsymbol{\mathrm{z}}|\boldsymbol{\mathrm{x}}_\mathrm{1}, \cdots, \boldsymbol{\mathrm{x}}_\mathrm{M}) \approx F(\{\boldsymbol{\mathrm{x}}_\mathrm{i}\}_{i=1}^M)$, where $F(\cdot)$ is a product or mean operation, depending on the chosen aggregation method. As discussed in \cite{sutter2021generalized}, these scalable multimodal VAE methods differ only in the choice of aggregation method. Different from those mentioned above multimodal VAE methods, we attain the joint posterior in its original form without introducing additional assumptions on the form of the joint posterior. To handle the issue of scalability, we exploit the deterministic set operation function in the noise-outsourcing process. While existing multimodal VAE methods can be viewed as typical late fusion method that combines decisions about the latent variables \cite{khaleghi2013multisensor}, the proposed SMVAE method corresponds to the early fusion method at the representation level, allowing for the learning of correlation and co-representation from multimodal data.

\subsection{Methods for set-input problems} 
Multiple instance learning (MIL) \cite{CARBONNEAU2018329} and 3D shape recognition \cite{su2015multi,hofer20053d,wu20153d}, are well-known examples of weakly-supervised learning problems that deal with set-input. MIL handles training data as numerous sets of instances with only set-level labels. A typical way to solve set-level classification problems is to use pooling methods for information aggregation \cite{shao2021transmil}. Recently, \cite{lee2019set} observed that classical feed-forward neural networks like the multi-layer perception (MLP) \cite{murtagh1991multilayer} cannot guarantee invariance under the permutation of the elements in the input as well as the input of arbitrary sizes. Furthermore, recursive neural networks such as RNN and LSTM \cite{hochreiter1997long} are sensitive to the order of the input sequences, and cannot fit the multimodal case since there is no natural order for modalities. Recently, Deep Sets \cite{zaheer2017deep} provided a formal definition for a permutation invariant function in set-input problems and proposed a universal approximator for arbitrary set functions. Later on, Set Transformer \cite{lee2019set} further extends this idea by using the self-attention mechanism to provide interactions as well as information aggregation among elements from an input set. However, their method only models a set of outputs as a deterministic function. Our work fills the gap between a deterministic set function to a probabilistic distribution and applies it to multimodal unsupervised learning. 
\section{Proposed Method}
\subsection{Preliminaries}\label{subsec:Preliminaries}
This work considers the multimodal learning problem as a set modeling problem and presents a scalable method for learning multimodal latent variables and cross-modality generation. Given a dataset $\{ \mathbb{X}^{(i)}\}^N_{i=1}$ of $N$ i.i.d. multimodal samples, we consider each of the sample as a set of $M$ modalities observations $\mathbb{X}^{(i)}= \{\boldsymbol{\mathrm{x}}_\mathrm{j}^{(i)}\}^M_{j=1}$. The multimodal data is assumed to be generated following the successive random process $p(\mathbb{X},\boldsymbol{\mathrm{z}})=p_\theta(\mathbb{X}|\boldsymbol{\mathrm{z}})p(\boldsymbol{\mathrm{z}}) $ which involves an unobserved latent variable $\boldsymbol{\mathrm{z}}$. The prior distribution of the latent variable $\boldsymbol{\mathrm{z}}$ is assumed to be $p_\theta(\boldsymbol{\mathrm{z}})$, with $\theta$ denoting its parameters. The marginal log-likelihood of this dataset of multimodal sets can be expressed as a summation of marginal log-likelihood of individual sets as $\log p(\mathbb{X}^{(i)})$ as $\log \prod^N_{i=1} p(\mathbb{X}^{(i)})=\sum^N_{i=1}\log p(\mathbb{X}^{(i)})$. Since the marginal likelihood of the dataset is intractable, we cannot optimize $p(\{\mathbb{X}^{(i)}\}^N_{i=1})$ with regards to $\theta$ directly. We instead introduce the variational approximation $q_{\phi}(\boldsymbol{\mathrm{z}}|\mathbb{X})$ from a parametric family, parameterized by $\phi$, as an importance distribution. $q_{\phi}(\boldsymbol{\mathrm{z}}|\mathbb{X})$ is often parameterized by a neural network with $\phi$ as its trainable parameters. Together, we can express the marginal log-likelihood of a single multimodal set as:
\begin{equation}\label{Set-ELBO}
\begin{split}
\log p(\mathbb{X}^{(i)}) &= D_{KL}(q_{\phi}(\boldsymbol{\mathrm{z}}|\mathbb{X}^{(i)}) || p_{\theta}(\boldsymbol{\mathrm{z}}|\mathbb{X}^{(i)})) + \mathcal{L}(\phi,\theta;\mathbb{X}^{(i)})\\
\mathcal{L}(\phi,\theta;\mathbb{X}^{(i)}) &= \mathbb{E}_{\boldsymbol{\mathrm{z}} \sim q_{\phi}(\boldsymbol{\mathrm{z}}|\mathbb{X}_{(i)})}\left[ \log p_{\theta}(\mathbb{X}^{(i)}),\boldsymbol{\mathrm{z}}) -\log q_{\phi}(\boldsymbol{\mathrm{z}}|\mathbb{X}^{(i)}) \right]\\
& = -D_{KL}(q_{\phi}(\boldsymbol{\mathrm{z}}|\mathbb{X}^{(i)}) || p_{\theta}(\boldsymbol{\mathrm{z}})) + \mathbb{E}_{\boldsymbol{\mathrm{z}} \sim q_{\phi}}(\boldsymbol{\mathrm{z}}|\mathbb{X}^{(i)}) \left[ \log p_{\theta}(\mathbb{X}^{(i)}|\boldsymbol{\mathrm{z}}) \right]
\end{split}
\end{equation}
, where $D_{KL}(\cdot||\cdot)$ is the Kullback-Leibler (KL) divergence between two distributions. The non-negative property of the KL divergence term between the variational approximation $q_{\phi}(\boldsymbol{\mathrm{z}}|\mathbb{X}^{(i)})$ and the true posterior $p_{\theta}(\boldsymbol{\mathrm{z}}|\mathbb{X}^{(i)})$ in the first line makes $\mathcal{L}(\phi,\theta;\mathbb{X}^{(i)})$ the natural evidence lower bound (ELBO) for the marginal log-likelihood. The last line indicates that maximizing the ELBO is equivalent to maximizing the reconstruction performance and regulating the variational approximation using the assumed prior distribution for the latent variable. To avoid confusion, we term neural networks used for mapping the raw input observations into a fixed-sized feature vector as the embedding network while the neural network used to parameterize the variational approximation $q_{\phi}(\boldsymbol{\mathrm{z}}|\mathbb{X}_{(i)})$ as the encoder network. A frequently used version of the objective function is written as:
\begin{equation}\label{Obj-SVAE}
\arg \underset{\phi} {min} -\beta D_{KL}( q_{\phi}(\boldsymbol{\mathrm{z}}|\mathbb{X}^{(i)}) || p(\boldsymbol{\mathrm{z}}) ) + \mathbb{E}_{\boldsymbol{\mathrm{z}} \sim q_{\phi}(\boldsymbol{\mathrm{z}}|\mathbb{X}^{(i)})}\left[ \lambda \log p(\mathbb{X}^{(i)}|\boldsymbol{\mathrm{z}}) \right] 
\end{equation}
, where additional annealing coefficients $\beta$ and reweighting coefficient $\lambda$ are used in the ELBO to allow gradients and warm-up training which gradually increases the regularization effect from the prior distribution and avoids reaching local minima in the early training stage \cite{bowman2015generating, sonderby2016ladder}. We drop the superscript of $\mathbb{X}^{(i)}$ to maintain brevity in the following paper. 

\subsection{Set Multimodal Variational Autoencoder}\label{subsec:4}
In multimodal scenarios with missing modalities, we consider each sample $\mathbb{X}_\mathrm{s}=\{\boldsymbol{\mathrm{x}}_\mathrm{i}|i^{th}$modaltiy present$\}$ as a subset of $\mathbb{X}$ and the powerset $\mathcal{P}(\mathbb{X})$ denoting all the $2^M$ combinations, such that $\mathbb{X}_\mathrm{s} \in \mathcal{P}(\mathbb{X})$. Our goal is to perform inference and generation from any number and permutation of available modalities, which requires an inference process is invariant to permutations and input of variable size. Following Definition \ref{d1}, we denotes the invariant inference process as $p(z|\mathbb{X}_\mathrm{s})=p(z|\pi{\cdot}\mathbb{X}_\mathrm{s})$. The ELBO for a subset $\mathbb{X}_\mathrm{s}$ can be written as Eq.\ref{ELBO-SMVAE}. 

\begin{equation}
\label{ELBO-SMVAE}
\mathcal{L}_s(\phi,\theta;\mathbb{X}_\mathrm{s})=-D_{KL}(q_{\phi}(\boldsymbol{\mathrm{z}}|\mathbb{X}_\mathrm{s}) || p_{\theta}(\boldsymbol{\mathrm{z}})) + \mathbb{E}_{\boldsymbol{\mathrm{z}} \sim q_{\phi}}(\boldsymbol{\mathrm{z}}|\mathbb{X}_\mathrm{s}) \left[ \log p_{\theta}(\mathbb{X}_\mathrm{s}|\boldsymbol{\mathrm{z}}) \right]
\end{equation}

\begin{definition}
\label{d1}
Let \(S_n\) be a set of all permutations of indices \({1,\cdots,N}\), \(X = (\boldsymbol{\mathrm{x}}_\mathrm{1}, \cdots \boldsymbol{\mathrm{x}}_\mathrm{n})\) denotes $n$ random variables. A probabilistic distribution \( p(y|X)\) is permutation inariant if and only if for any permutation \(\pi \in S_n\),  \(p(y| X)=p(y|\pi{\cdot}X)\), where \( \cdot \) is the group action.
\end{definition}

The difference between $\mathcal{L}(\phi,\theta;\mathbb{X})$ in Eq.\ref{Set-ELBO} and $\mathcal{L}_s(\phi,\theta;\mathbb{X}_\mathrm{s})$ in Eq.\ref{ELBO-SMVAE} is that the ELBO for a subset $\mathbb{X}_\mathrm{s}$ is not yet a valid bound for $\log p(\mathbb{X})$ by itself. Additional sampling from $\mathcal{P}(\mathbb{X})$ in the optimization objective as Eq.\ref{Obj-SMVAE} is needed for theoretical completeness. 
\begin{equation}
\label{Obj-SMVAE}
\arg \underset{\phi} {\min} \sum_{\mathbb{X}_\mathrm{s} \sim \mathcal{P}(\mathbb{X}) \atop \pi \in S_n } \mathcal{L}_s(\phi,\theta;\pi{\cdot}\mathbb{X}_\mathrm{s})
\end{equation}
, where $\pi$ is a randomly generated permutation to the input subset $\mathbb{X}_\mathrm{s}$. However, this sampling process can be trivial if we combine the sampling of the subsets with the sampling of mini-batch during training. By assuming the Gaussian form of the latent variable $\boldsymbol{\mathrm{z}}$ and applying the reparameterization technique, the inference process of SMVAE can be written as: 
\begin{equation}
\label{eq:Inference}
p(\boldsymbol{\mathrm{z}} | \boldsymbol{\mathrm{x}}_\mathrm{s} )\sim \mathcal{N}(\mu, \sigma^2), \epsilon \sim \mathcal{N}(0,I)
\end{equation}
\begin{equation}
\label{eq:noise-outsourced}
\boldsymbol{\mathrm{z}} := \mu + \sigma \odot \epsilon
\end{equation}
\begin{equation}
\label{eq:opt}
\mu_z, \log\sigma_z^2 := g_{\phi}(E_1(\boldsymbol{\mathrm{x}}_\mathrm{1}), \cdots, E_m(\boldsymbol{\mathrm{x}}_\mathrm{m}))
\end{equation}
, where $E_i$ are embedding network for the $i^{th}$ modality, $g_{\phi}(\cdot)$ is a neural network with trainable parameters $\phi$ that provide the parameter for the latent's posterior distribution (i.e., $\mu$ and $\sigma$) , $\odot$ denotes the element-wise multiplication. For the generation process, it is desired to models the joint likelihood of modalities conditioned on the latent variables $p_\theta(\boldsymbol{\mathrm{x}}_\mathrm{s},\boldsymbol{\mathrm{z}})=p(\boldsymbol{\mathrm{z}})p_{\theta}(\boldsymbol{\mathrm{x}}_\mathrm{s}|\boldsymbol{\mathrm{z}})$ so that the model can utilize information from other available modalities more easier when generating a complex modality. However, for the sake of easy implementation, we assign $n$ separate decoders $D_1,\cdots, D_M$ for all possible modalities as $p_\theta(\boldsymbol{\mathrm{x}}_\mathrm{s}|\boldsymbol{\mathrm{z}}) = \left[D_{\theta_1}(z), \cdots, D_{\theta_M}(z) \right]$. We find empirically that, without loss of generality, using $L_2-$normalization as additional regularization to regulate the parameter o$\mu$ and $\sigma$ of the inference network to $0$ and $1$ respectively could facilitate the learning efficiency because the gradient from the ELBO often favors the reconstruction term over the regularization term. 

\subsection{Set Representation for Joint Distribution}\label{subsec:attention}
The scalability issue comes from the requirement for an inference process for the powerset $\mathcal{P}(\mathbb{X})$. We achieve scalability by using the noise-outsourced functional representation, i.e. $\boldsymbol{\mathrm{z}}=g(\epsilon,\mathbb{X}_\mathrm{s})$, to bridge the gap between the deterministic set functions to a stochastic function. The properties of the deterministic function thus can be passed to the stochastic distribution under minor conditions \cite{bloem2020probabilistic}. With such a foundation, the problem of modeling the posterior for a superset immediately reduces to designing a differentiable deterministic function that has the desired invariant or elastic properties. Specifically, we identify four critical requirements for weakly-supervised multimodal learning. Being that the model should 1) be scalable in the number of observable modalities; 2) be able to process input modalities sets of arbitrary size and permutation; 3) satisfy Theorem \ref{theorem1}; and 4) be able to learn the co-representation among all modalities. 
\begin{theorem}\label{theorem1}
A valid set function \(f(x)\) is invariant to the permutation of instances, \textit{iif} it can be decomposed in the form \(\Phi(\sum \Psi(x)) \), for any suitable transformations \(\Phi\) and \(\Psi\).
\end{theorem}

An oversimplified example of a set function can be summation or product as done in MVAE \cite{wu2018multimodal} and MMVAE \cite{shi2019variational}. Pooling operations such as average pooling or max pooling also fit the definition. However, these set aggregation operations will require additional factorization assumptions to the joint posterior and ultimately forbid the VAE to learn co-representation of the input modalities as aggregation is only applied at the decision level. To establish the inductive bias of inter-modality correlation, the self-attention mechanism without positional embeddings is a reasonable choice \cite{edelman2022inductive,shvetsova2022everything}. 

Therefore, the proposed SMVAE leverages self-attention as the deterministic set function to aggregate embeddings of multimodal inputs. Given the query $Q$, key $K$ and value $V$, an attention function is denoted as $\mathrm{Att}(Q,K,V)=\omega(\frac{QK^T}{\sqrt{d_k}})V$, where $K \in \mathbb{R}^{m \times d_k}$ and $V \in \mathbb{R}^{m \times d_v}$ are $m$ vectors of dimension $d_k$ and $d_v$, $Q \in \mathbb{R}^{n \times d_q}$ are $n$ vectors of dimension $d_q$, $\omega$ is the softmax activation function. In our case, the key-value pairs represent the $m$ available embeddings of input modalities, $m \le M$. Each embedding is mapped to a $d-$dimensional embedding space by a modality-specific embedding network. By measuring the compatibility of the corresponding key and the query $Q$, information that is shared among modalities is aggregated as co-representation.


In practice, we utilize the multi-head extension of self-attention denoted as $\mathrm{MultiHead}(Q,K,V,h)=\mathrm{Concat}(A_1,\cdots,A_h)W^o$, where $A_i = \mathrm{Att}_i(QW_i^Q,KW_i^K,VW_i^v)$ is obtained from the $i^{th}$ attention function with projection parameters $W_i^Q \in \mathbb{R}^{(d/h)\times{d_q}}$,$W_i^K \in \mathbb{R}^{(d/h)\times{d_k}}$, $W_i^V \in \mathbb{R}^{(d/h)\times{d_k}}$ and $W^o \in \mathbb{R}^{d_v\times{d}}$, $h$ denotes the total number of attention heads and $d$ denotes the dimension of the projections for keys, values and queries. Inspired by \cite{lee2019set}, we design our deterministic set representation function $g_{\phi}(\mathbb{X}_\mathrm{s})$ as follows:
\begin{equation}\label{Set-attention}
\begin{split}
g_{\phi}(\mathbb{X}_\mathrm{s}) := H + f_s(H) &\\
H = I + \mathrm{MultiHead}(I,\mathbb{X}_\mathrm{s},\mathbb{X}_\mathrm{s},h)&
\end{split}
\end{equation}
, where $I \in R^{1 \times d_v}$ is an $d_v$-dimensional trainable vector as the query vector for multimodal embeddings. $f_s$ is a fully-connected layer. By calculating attention weights using $I$ and each embedding. Not only does $I$ work as an aggregation vector that regulates the number of output vectors from $g_{\phi}(\mathbb{X}_\mathrm{s})$ to be constant regardless of the number of input embeddings, but also it selects relevant information from each embedding base on similarity measurement. The former justifies $g_{\phi}(\mathbb{X}_\mathrm{s})$ as a suitable permutation invariant set-processing function while the latter yields the desired co-representation among modalities. Finally, Since the set representation function $g_{\phi}(\mathbb{X}_\mathrm{s})$ is invariant to the input permutations of different input sizes, we achieved an invariant inference probabilistic function that satisfies Definition \ref{d1} through the noise-outsourced process as shown in Eq. \ref{eq:noise-outsourced}. Thus, by introducing the set representation function in the noise-outsourced process, the SMVAE is readily a scalable multimodal model for any subsets of modalities.

\subsection{Total Correlation Optimization without Condition Independence}\label{subsec:5}
The lower bound of the multimodal data without factorizing the joint posterior (i.e., Eq. \ref{Set-ELBO}) provides additional information about the correlations of modalities during the optimization process compared to factorized methods. It is noteworthy that both MVAE and MMVAE depend on the assumption of conditional independence between modalities in factorization. Without loss of generality, the relation between $\mathcal{L}(\phi,\theta;\mathbb{X})$ and the factorized case $\mathcal{L}_{C\!I\!}$ can be shown in Eq. \ref{ELBO-CI}.
\begin{equation}
    \label{ELBO-CI}
    \begin{split}
    \mathcal{L}(\phi,\theta;\mathbb{X}) &=\mathbb{E}_{q_\phi(\mathbf{z}\mid\mathbb{X})} \left[\log \frac{p_\theta(z)\prod_{i=1}^Mp_\theta(x_i\mid z)}{q_\phi(\mathbf{z}\mid\mathbb{X})}+ \log \frac{p_\theta(\mathbb{X}, \mathbf{z})}{p_\theta(z)\prod_{i=1}^Mp(x_i\mid z)}\right] \\
    &=\mathcal{L}_{C\!I\!} + \mathbb{E}_{q_\phi(\mathbf{z}\mid\mathbb{X})} \left[\log \frac{p_\theta(\mathbb{X}\mid \mathbf{z})}{\prod_{i=1}^Mp_\theta(\boldsymbol{\mathrm{x}}_i\mid z)} \right]
    \end{split}
\end{equation}
, where $\mathbb{X} \equiv (\boldsymbol{\mathrm{x}}_1,\cdots,\boldsymbol{\mathrm{x}}_M)$ and $\mathcal{L}_{C\!I\!}$ is the lower bound for factorizable generative process as MVAE or MMVAE. Specifically, let $q(\mathbb{X})$ denotes the empirical distribution for the multimodal dataset, we have:
\begin{equation}
    \label{CTC}
    \mathbb{E}_{q(\mathbb{X})}\left[\mathcal{L}(\phi,\theta;\mathbb{X})\right]=\mathbb{E}_{q(\mathbb{X})}\left[\mathcal{L}_{CI}\right]+ \mathbb{E}_{z\sim \frac{q(\mathbb{X}) q_\phi(z\mid\mathbb{X}) }{p_\theta(\mathbb{X}\mid z)}} \left[ \underset{\text{conditional total correlation}}{\underbrace{\mathbb{E}_{\mathbb{X}\sim p_\theta(\mathbb{X}\mid \mathbf{z})}\left[\log \frac{p_\theta(\mathbb{X}\mid \mathbf{z})}{\prod_{i=1}^Mp_\theta(x_i\mid \mathbf{z})}
\right]}}\right]
\end{equation}
, which reveals that without assuming a factorizable generative process and enforcing conditional independence among modalities, our optimization objective naturally models the conditional total correlation which provides information of dependency among multiple input modalities \cite{watanabe1960information,studeny1998multiinformation}. Therefore, the SMVAE has the additional advantage of learning correlations among different modalities of the same event, which is also what we desired for good co-representation. 

\section{Experiments}\label{Experiments}
\subsection{Experiment settings} 
We make use of uni-modal datasets including MNIST \cite{lecun1998gradient}, FashionMNIST \cite{xiao2017online} and CelebA \cite{liu2015faceattributes} to evaluate the performance of the proposed SMVAE and compare with other state-of-the-art methods. We convert these uni-modal datasets into bi-modal dataset by transforming the labels to one-hot vectors as the second modality as in \cite{wu2018multimodal,suzuki2016joint}. For quatitative evaluation, we denote $\boldsymbol{\mathrm{x}}_\mathrm{1}$ and $\boldsymbol{\mathrm{x}}_\mathrm{2}$ as the image and text modality and measure the marginal log-likelihood, $\log p(\mathbf{x}) \approx \log E_{q(\mathbf{z}|\cdot)}[\frac{p(\mathbf{x}|\mathbf{z})p(\mathbf{z})}{q(\mathbf{z}|\cdot)}]$, the joint likelihood $\log p(\mathbf{x},\mathbf{y}) \approx \log E_{q(\mathbf{z}|\cdot)}[\frac{p(\mathbf{z})p(\mathbf{x}|\mathbf{z})p(\mathbf{y}|\mathbf{z})}{q(\mathbf{z}|\cdot)}] $, and the marginal conditional probability, $\log p(\mathbf{x}|\mathbf{y})\approx\log E_{q(\mathbf{z}|\cdot)}[\frac{p(\mathbf{z})p(\mathbf{x}|\mathbf{z})p(\mathbf{y}|\mathbf{z})}{q(\mathbf{z}|\cdot)}]-\log E_{p(\mathbf{z})}[p(\mathbf{y}|\mathbf{z})]$, using data samples from the test set. $q(\mathbf{z}|\cdot)$ denotes the importance distribution. For all the multimodal VAE methods, we keep the architecture of encoders and decoders consistent for a fair comparison. Detailed training configurations and settings of the networks are listed in Appendix. \ref{Appendix-Arch}. The marginal probabilities measure the model's ability to capture data distributions while the conditional log probability measures classification performance. Higher scoring of these matrics means the better a model is able to generate proper samples and convert between modalities. These are the desirable properties for learning a generative model. 

\begin{figure}[!hb]
        \vspace{-0.5mm}
        \centering        
        \setlength{\abovecaptionskip}{0.cm}
        \subfigure[MVAE]{\includegraphics[scale=0.1]{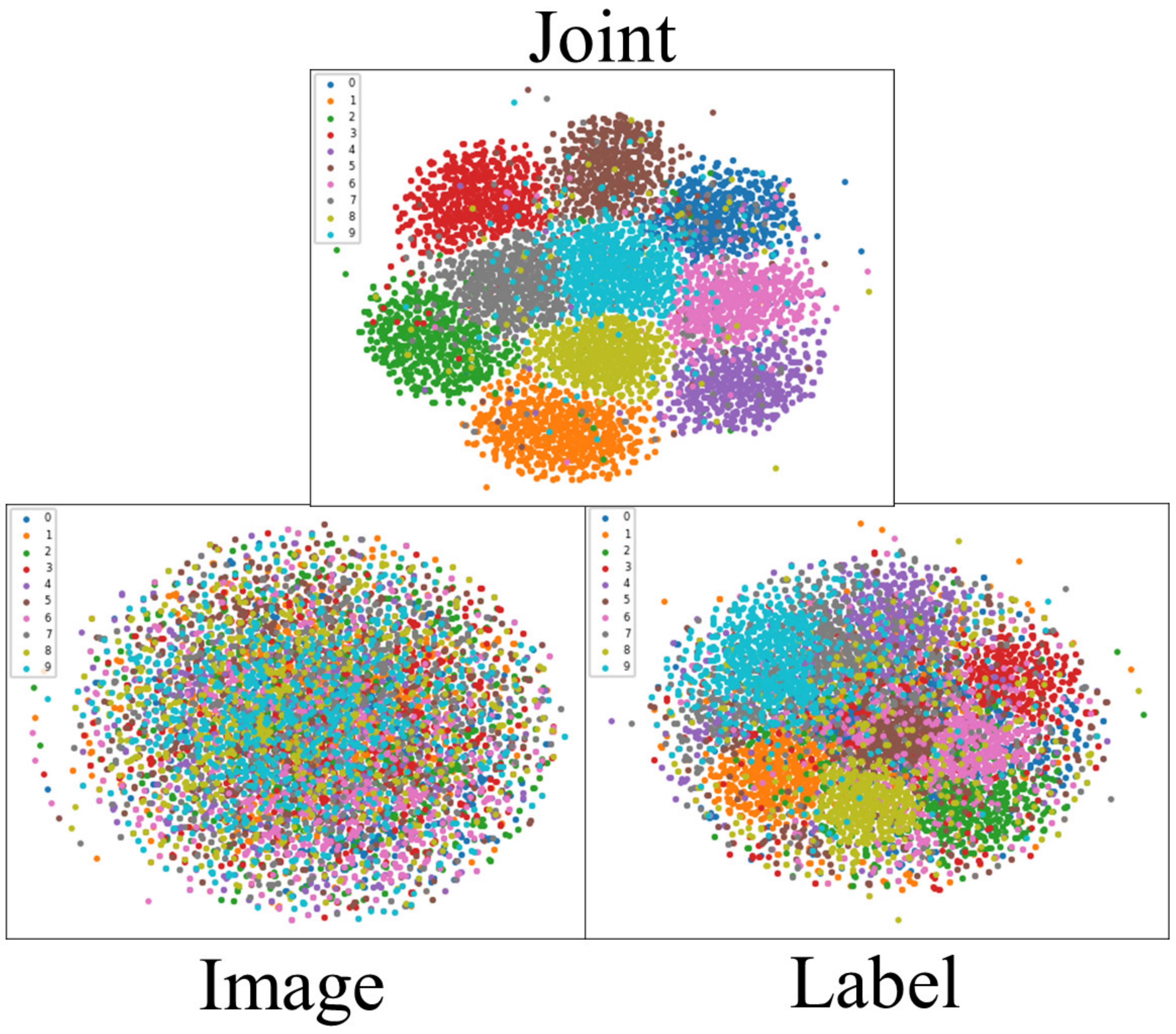}}
        \subfigure[SMVAE]{\includegraphics[scale=0.1]{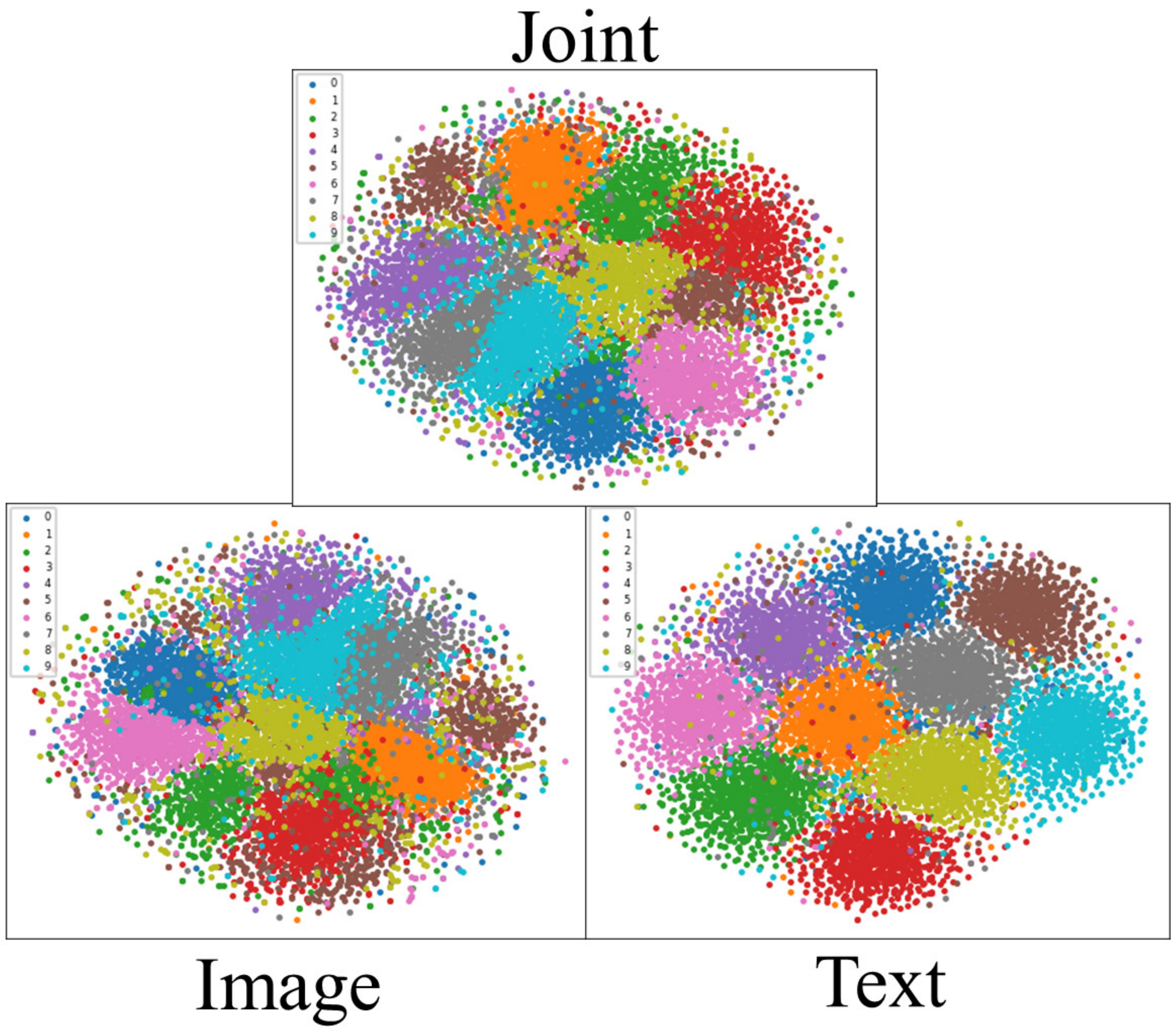}}
        \subfigure[MMVAE]{\includegraphics[scale=0.1]{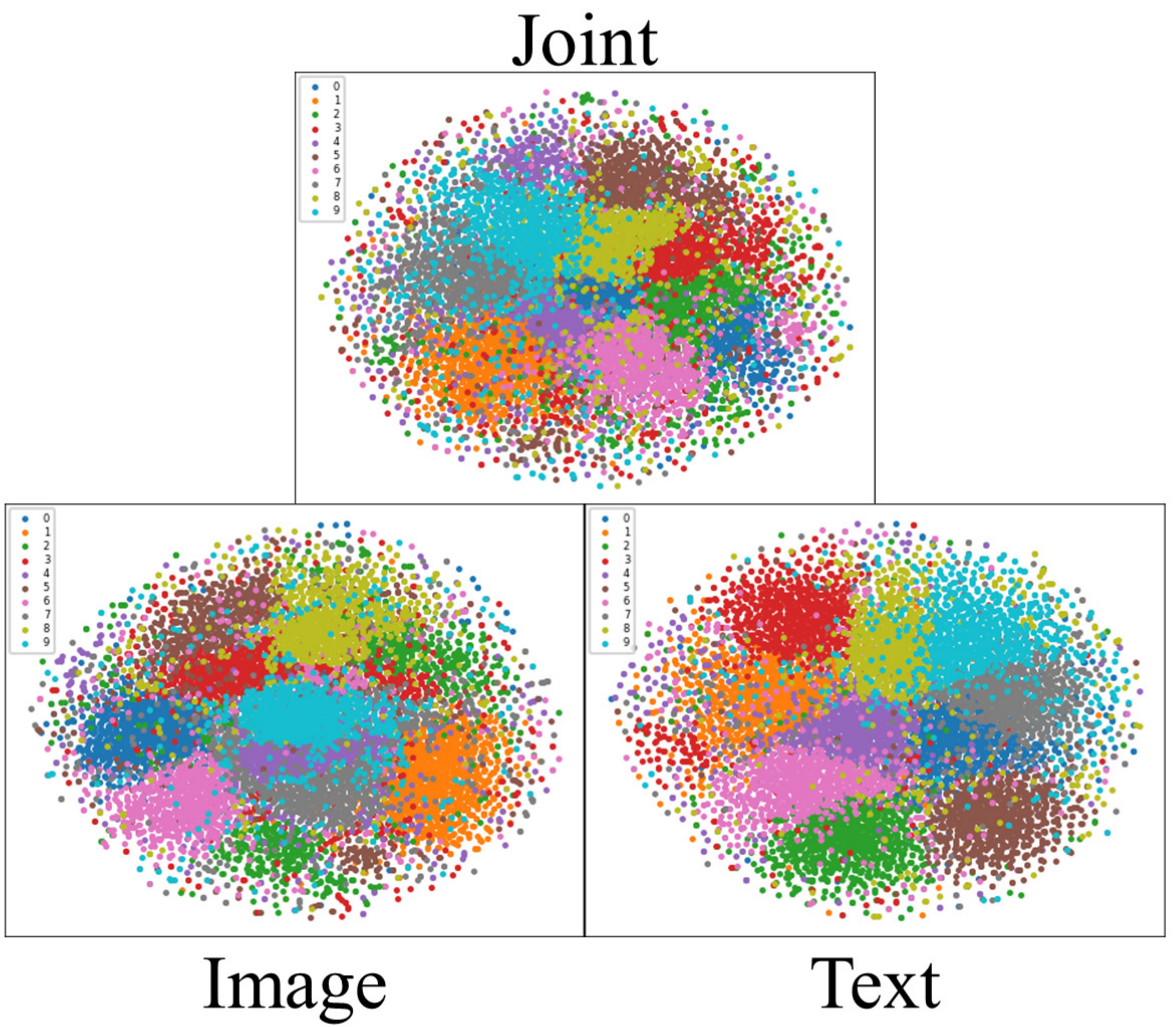}}
        \caption{Visualization of $2$-D latent representation learned by our model compared to MVAE and MMVAE from bi-modality MNIST dataset. Our model can learn natrual clusters while maintaining good structure for their joint distribution. Left: MVAE. Middle: SMVAE. Right: MMVAE.\label{fig:ClatentV}}
\end{figure}
\subsection{Generation quality and Quantitative evaluation}
\begin{figure}[!ht]
  \setlength{\abovecaptionskip}{0.2 cm}
  \centering
  \subfigure[digit 0]{\includegraphics[scale=0.28]{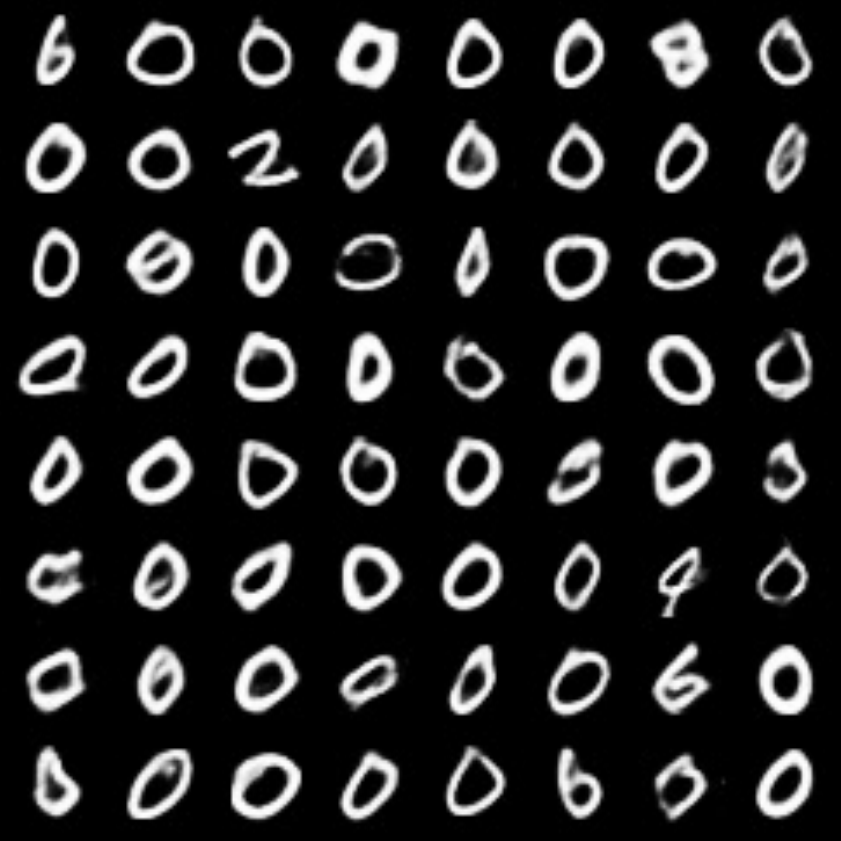}}\vspace{-0.5mm}
  \subfigure[digit 1]{\includegraphics[scale=0.28]{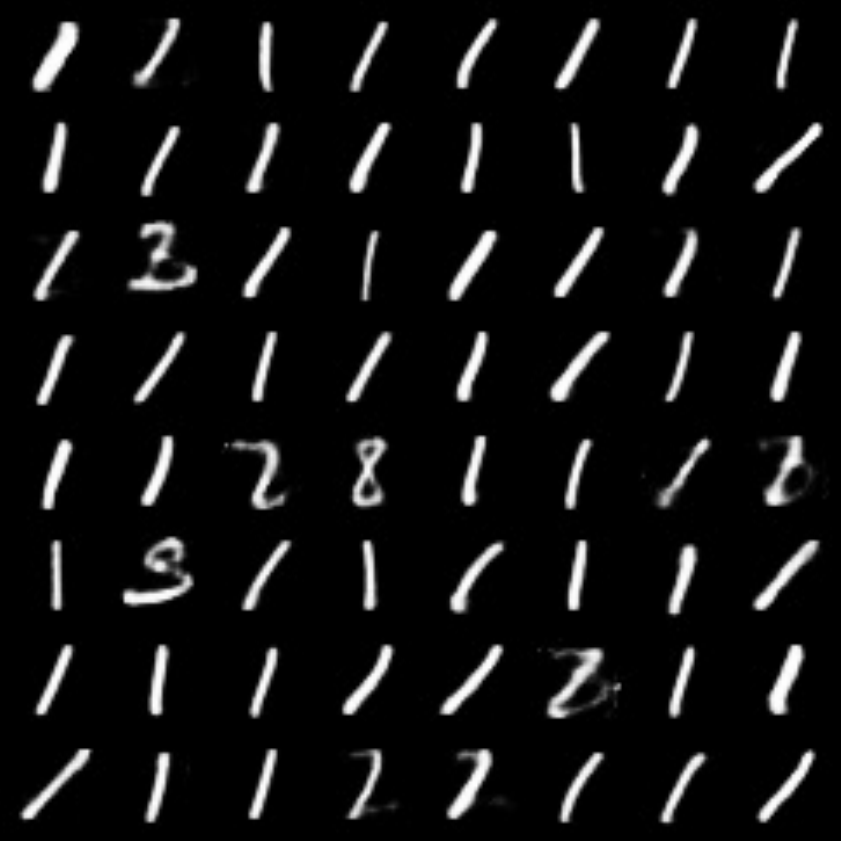}}\vspace{-0.5mm}
  \subfigure[digit 2]{\includegraphics[scale=0.28]{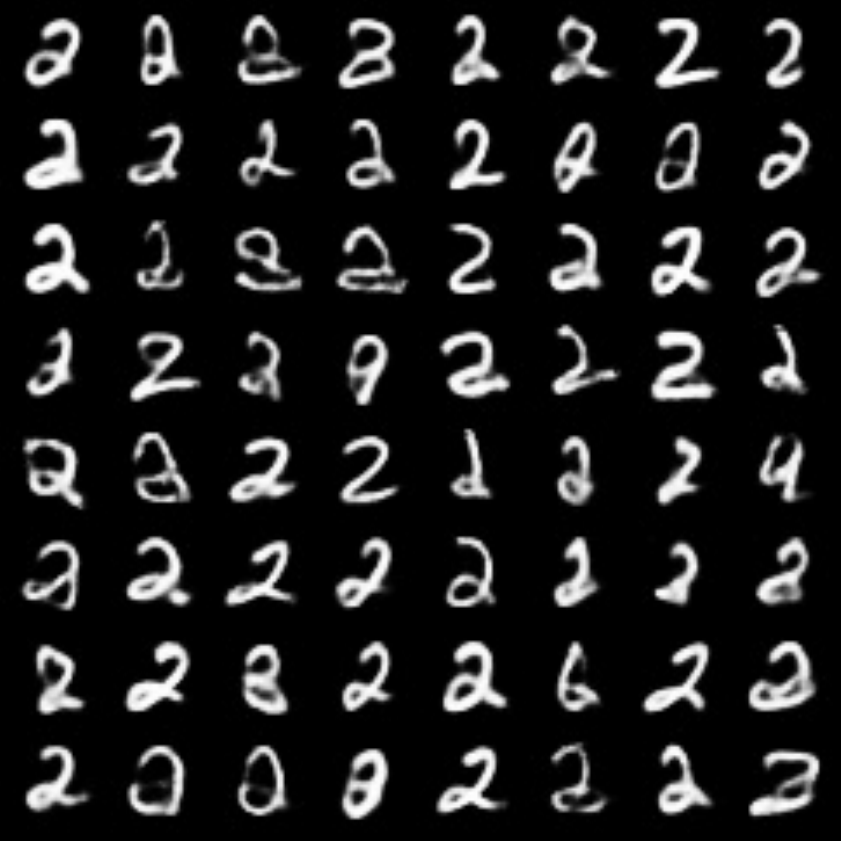}}\vspace{-0.5mm}
  \subfigure[digit 3]{\includegraphics[scale=0.28]{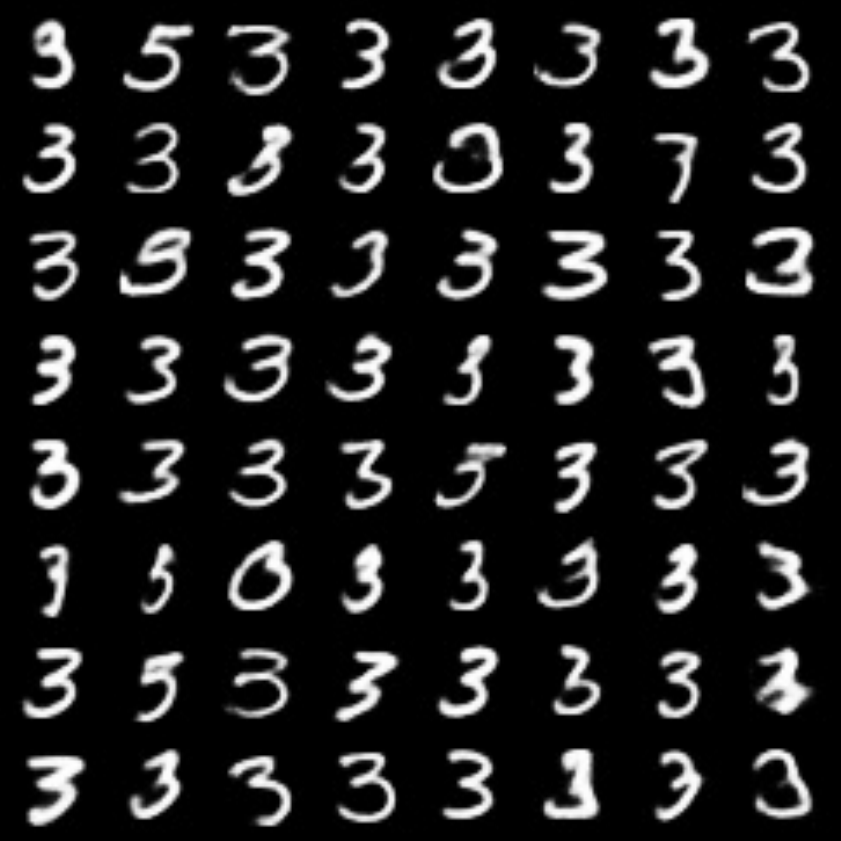}}\vspace{-0.5mm}
  \subfigure[digit 4]{\includegraphics[scale=0.28]{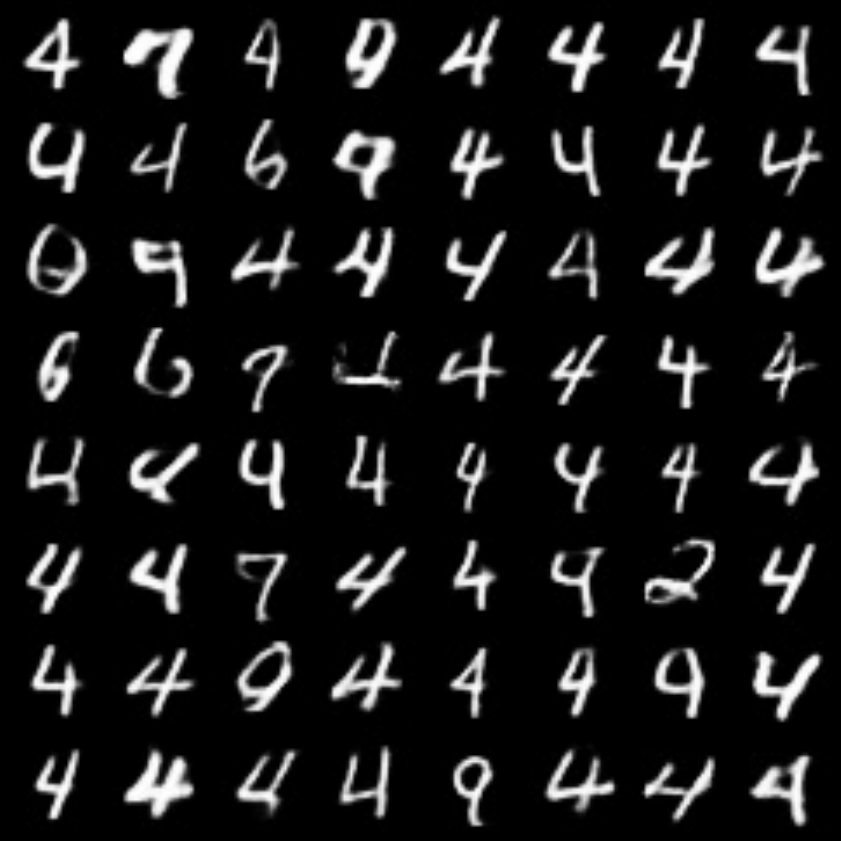}}\vspace{-0.5mm}
  \subfigure[digit 5]{\includegraphics[scale=0.28]{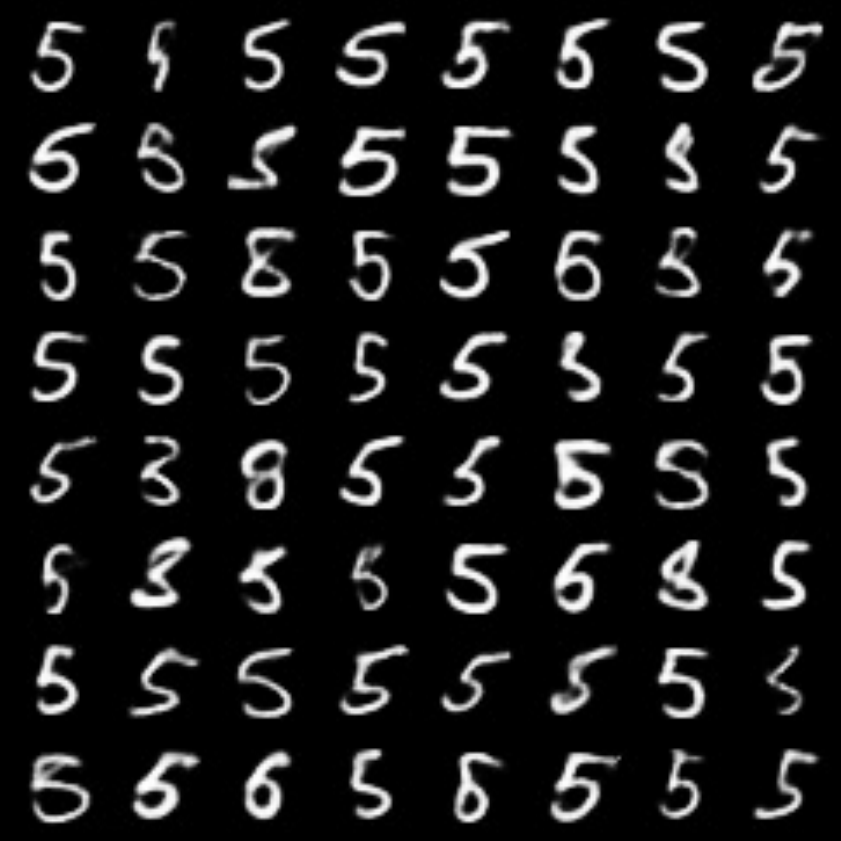}}\vspace{-0.5mm}
  \subfigure[digit 6]{\includegraphics[scale=0.28]{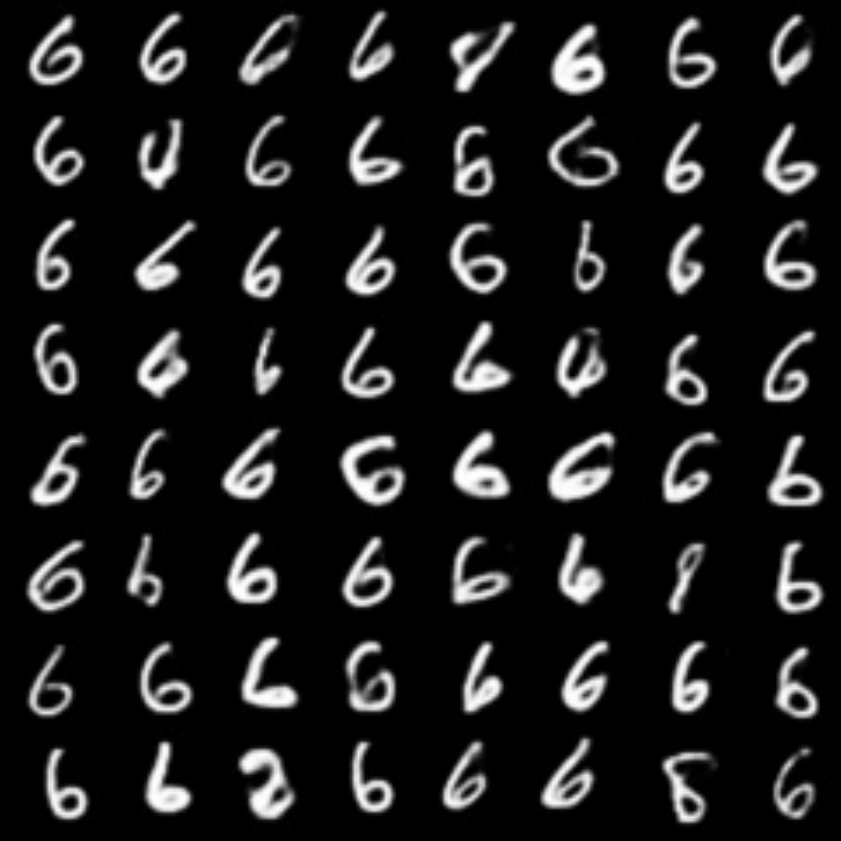}}\vspace{-0.5mm}
  \subfigure[digit 7]{\includegraphics[scale=0.28]{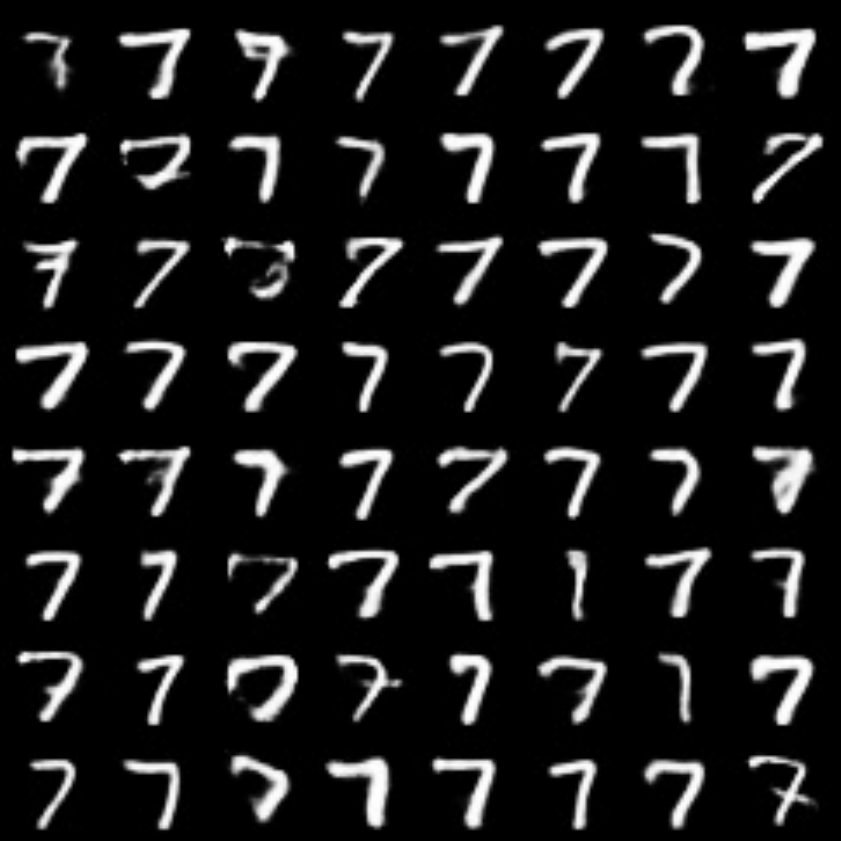}}\vspace{-0.5mm}
  \subfigure[digit 8]{\includegraphics[scale=0.28]{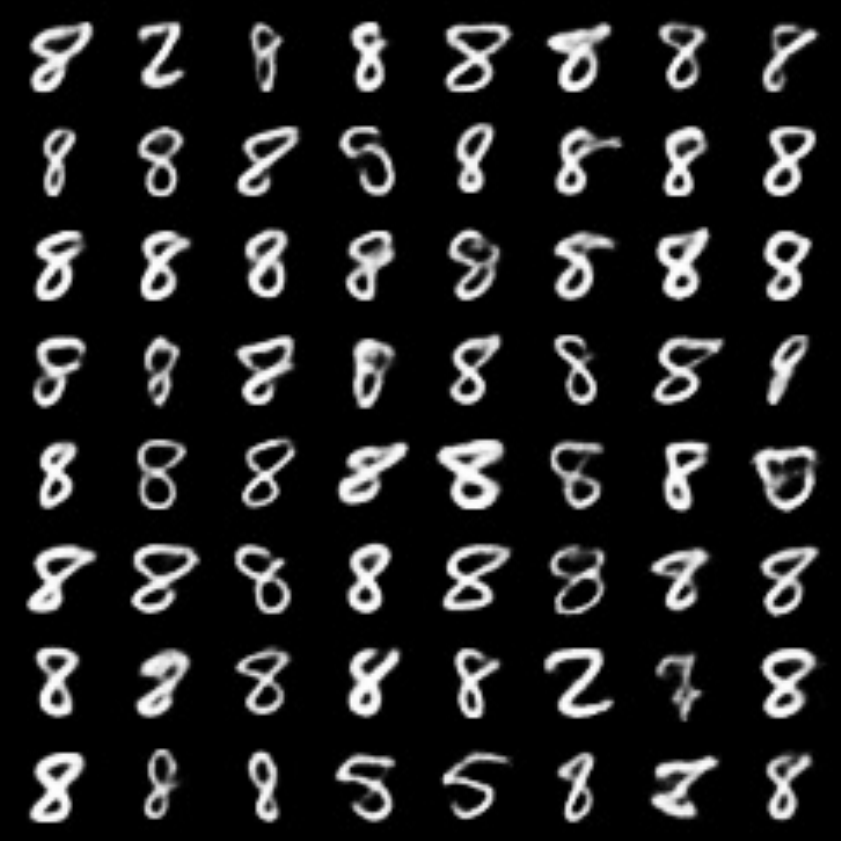}}\vspace{-0.5mm}
  \subfigure[digit 9]{\includegraphics[scale=0.28]{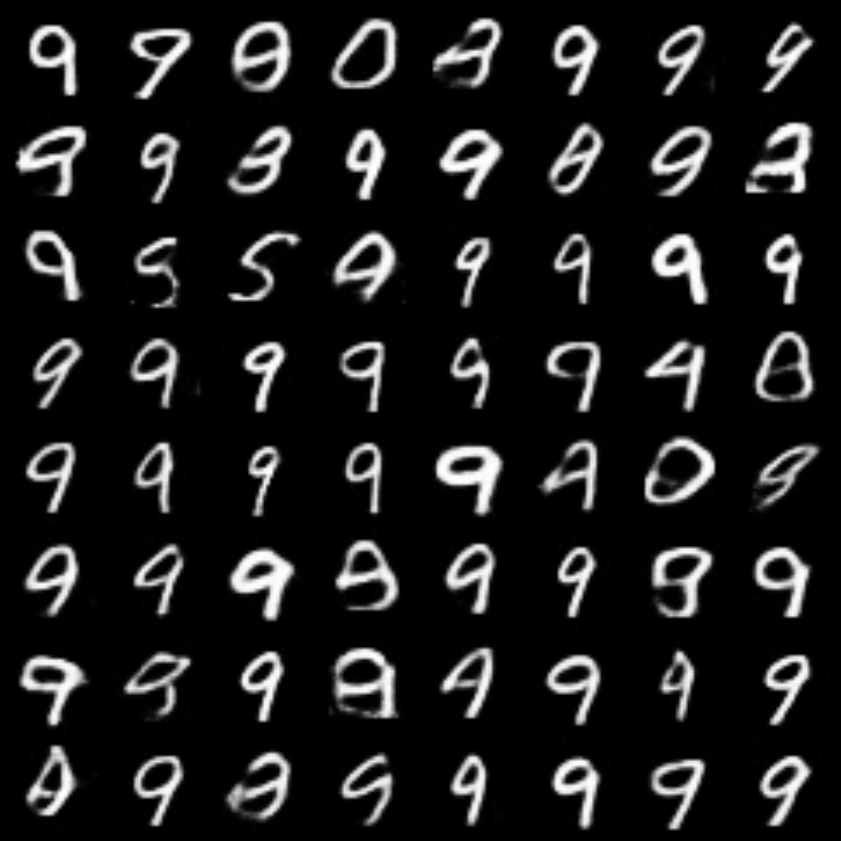}}\vspace{-0.5mm}
  \caption{Conditional image reconstruction of digits generated by SMVAE given text input.\label{fig:mnist}}
\end{figure}

\begin{figure}[!ht]
  \setlength{\abovecaptionskip}{0.2 cm}
  \centering
  \subfigure[Ankle boot]{\includegraphics[scale=0.28]{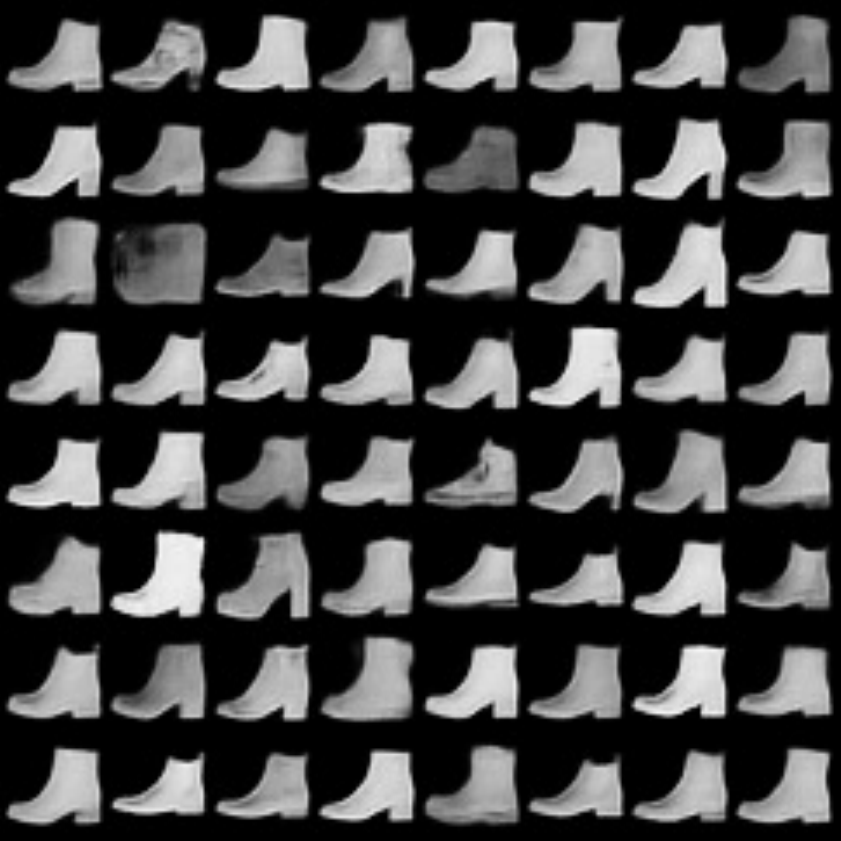}}\vspace{-0.5mm}
  \subfigure[Bag]{\includegraphics[scale=0.28]{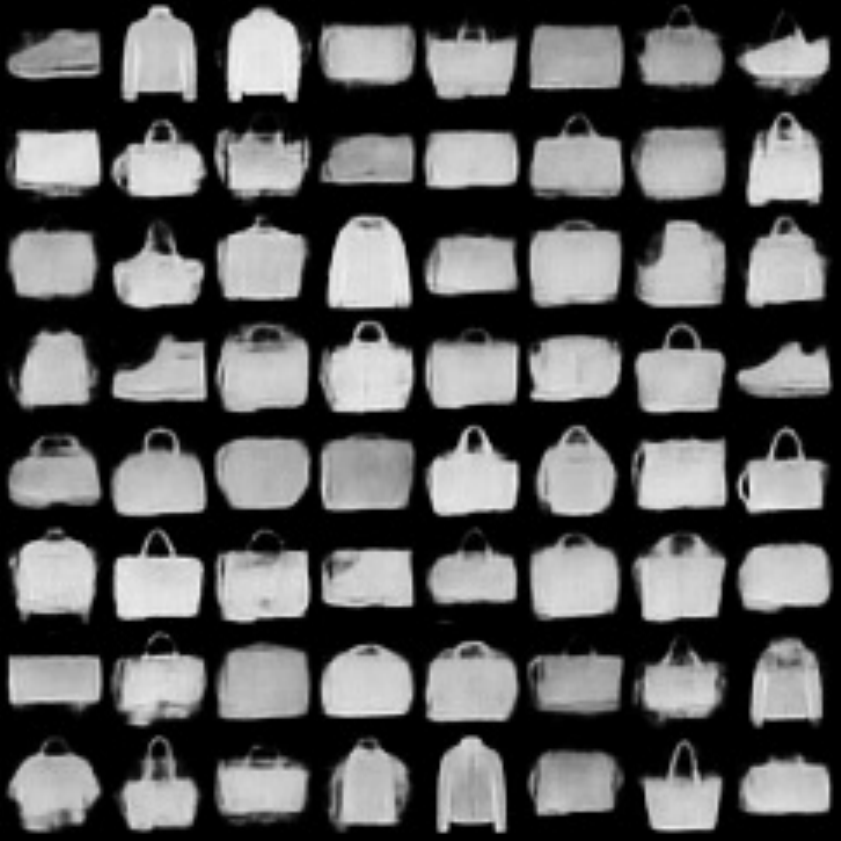}}\vspace{-0.5mm}
  \subfigure[Coat]{\includegraphics[scale=0.28]{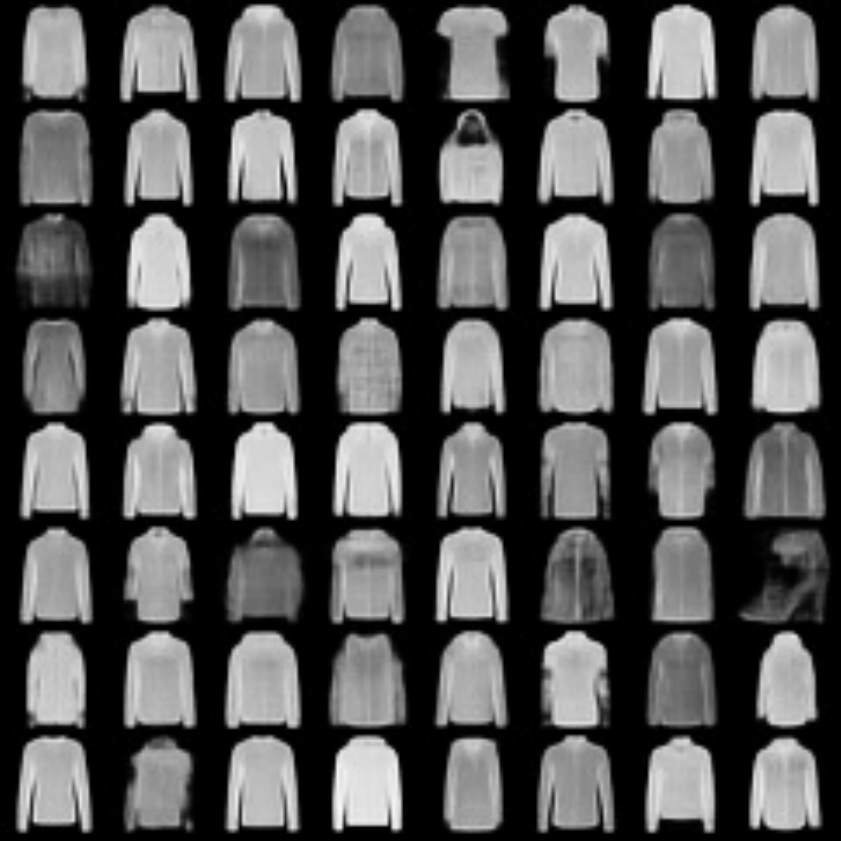}}\vspace{-0.5mm}
  \subfigure[Dress]{\includegraphics[scale=0.28]{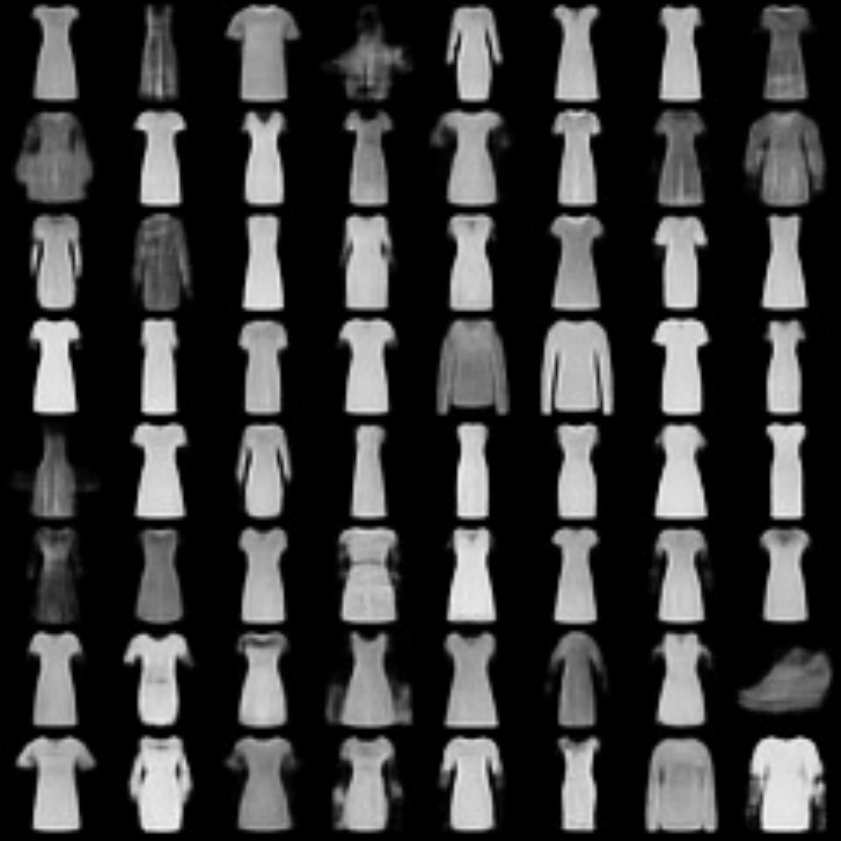}}\vspace{-0.5mm}
  \subfigure[Pullover]{\includegraphics[scale=0.28]{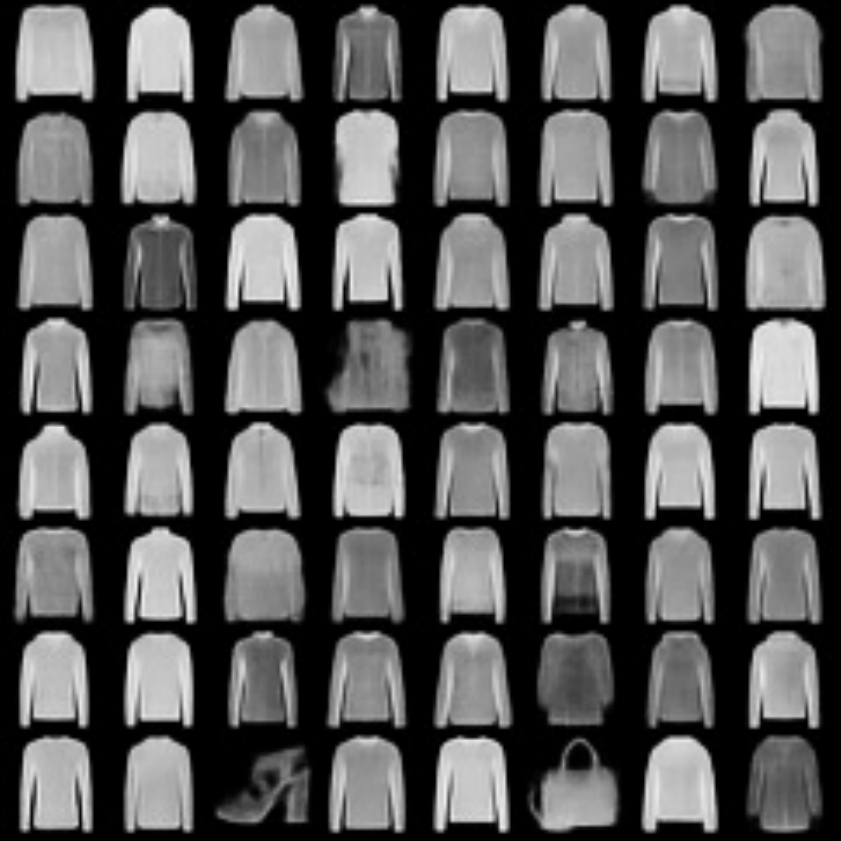}}\vspace{-0.5mm}
  \subfigure[Sandal]{\includegraphics[scale=0.28]{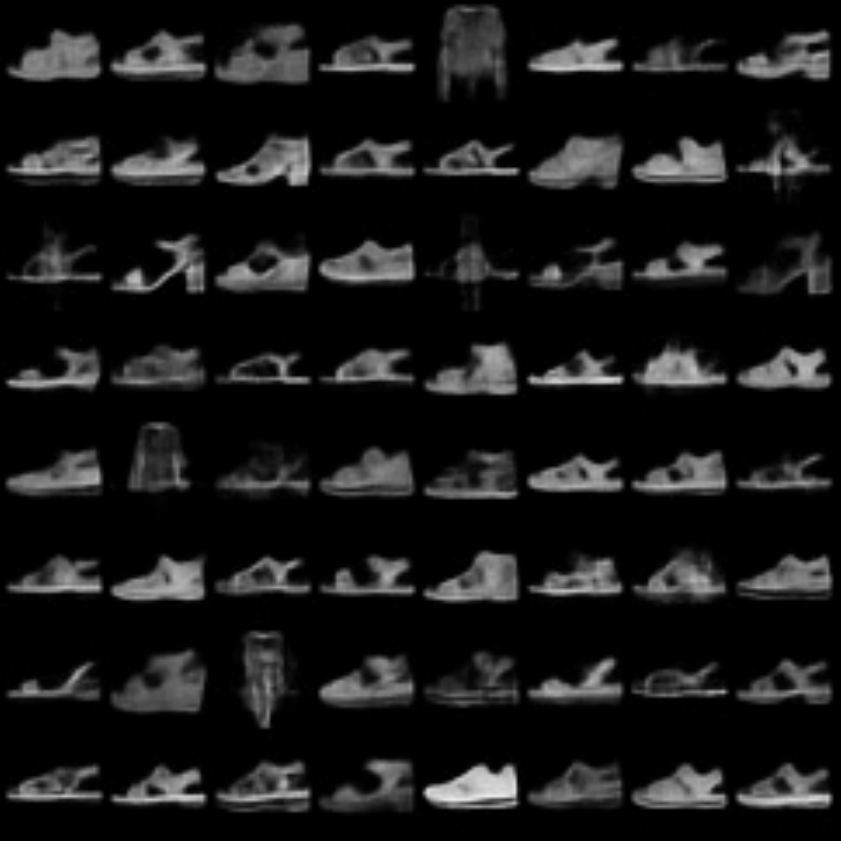}}\vspace{-0.5mm}
  \subfigure[Shirt]{\includegraphics[scale=0.28]{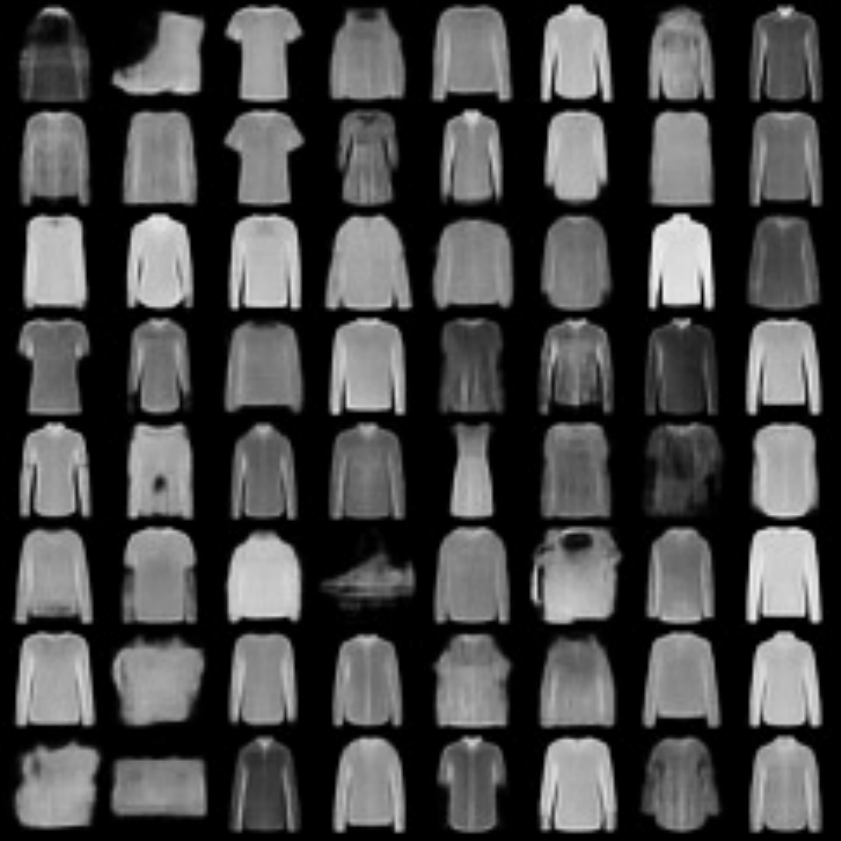}}\vspace{-0.5mm}
  \subfigure[Sneaker]{\includegraphics[scale=0.28]{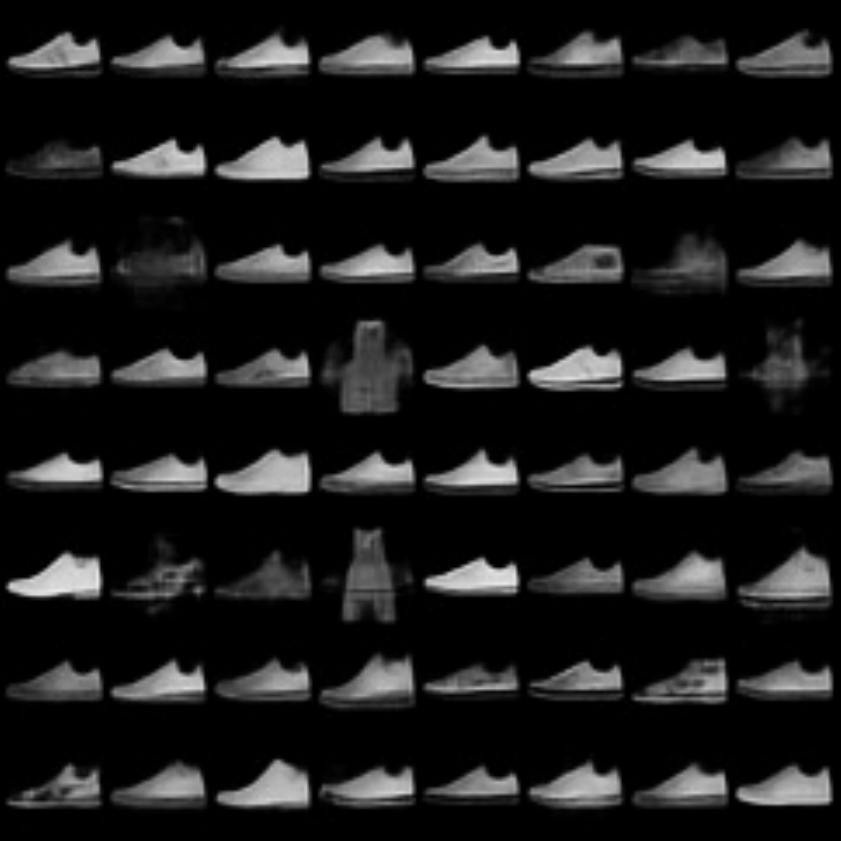}}\vspace{-0.5mm}
  \subfigure[Trouser]{\includegraphics[scale=0.28]{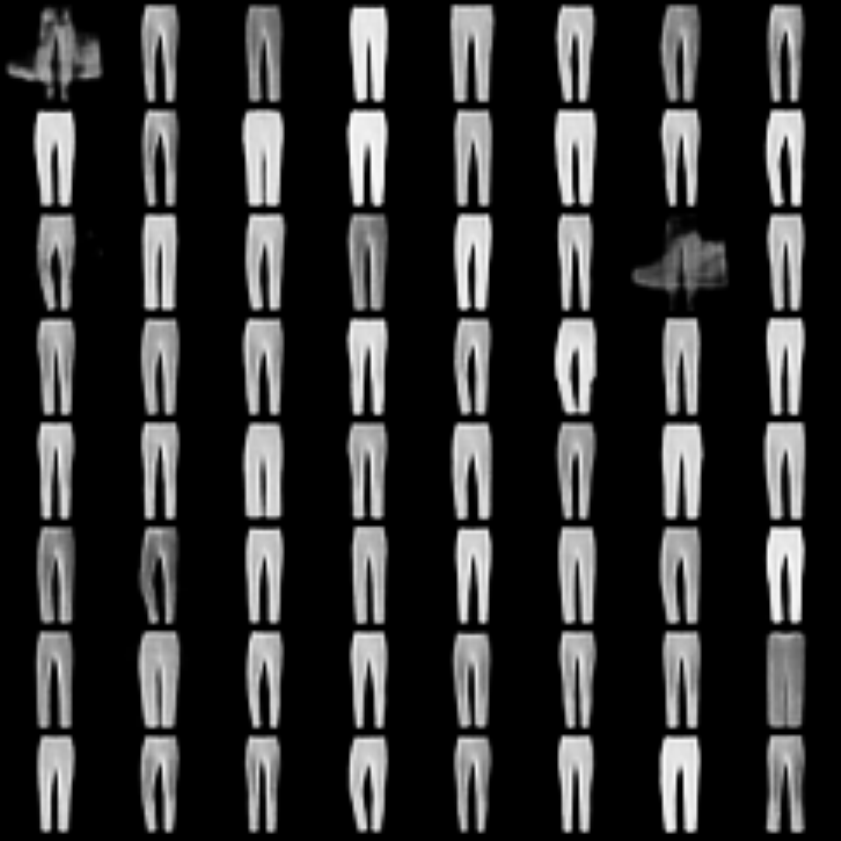}}\vspace{-0.5mm}
  \subfigure[T-shirt]{\includegraphics[scale=0.28]{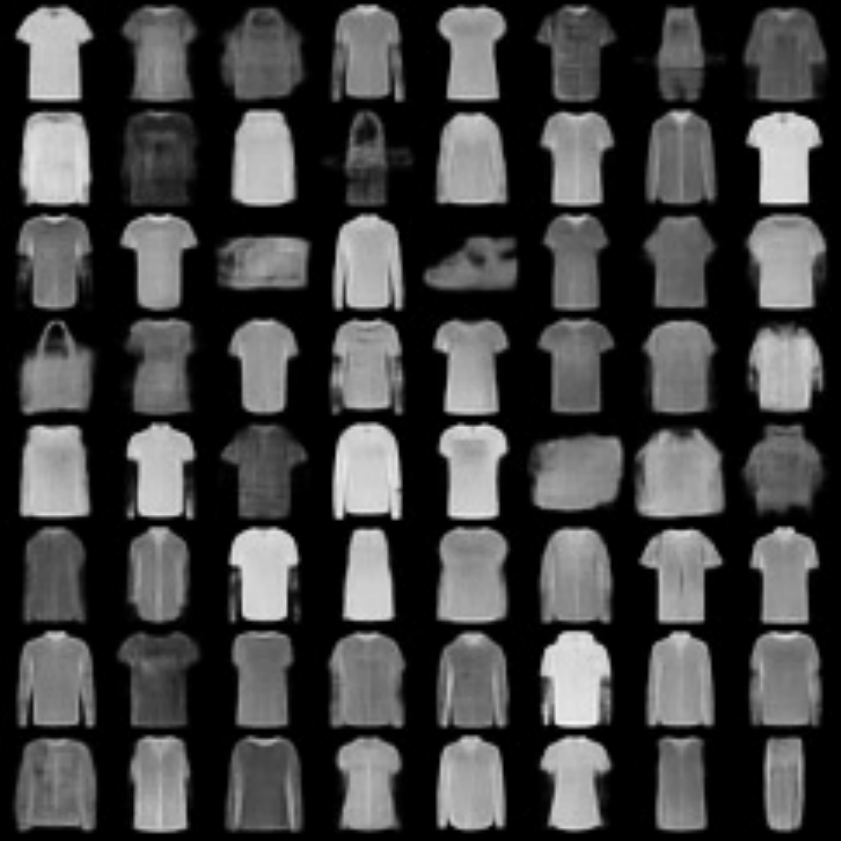}}\vspace{-0.5mm}
  \caption{Conditional image reconstruction of fashionMNIST generated by SMVAE given text input.\label{fig:fashion}}
\end{figure}

\begin{figure}[!ht]
 \setlength{\abovecaptionskip}{0.cm}
  \centering
  \subfigure[Black Hair]{\includegraphics[scale=0.25]{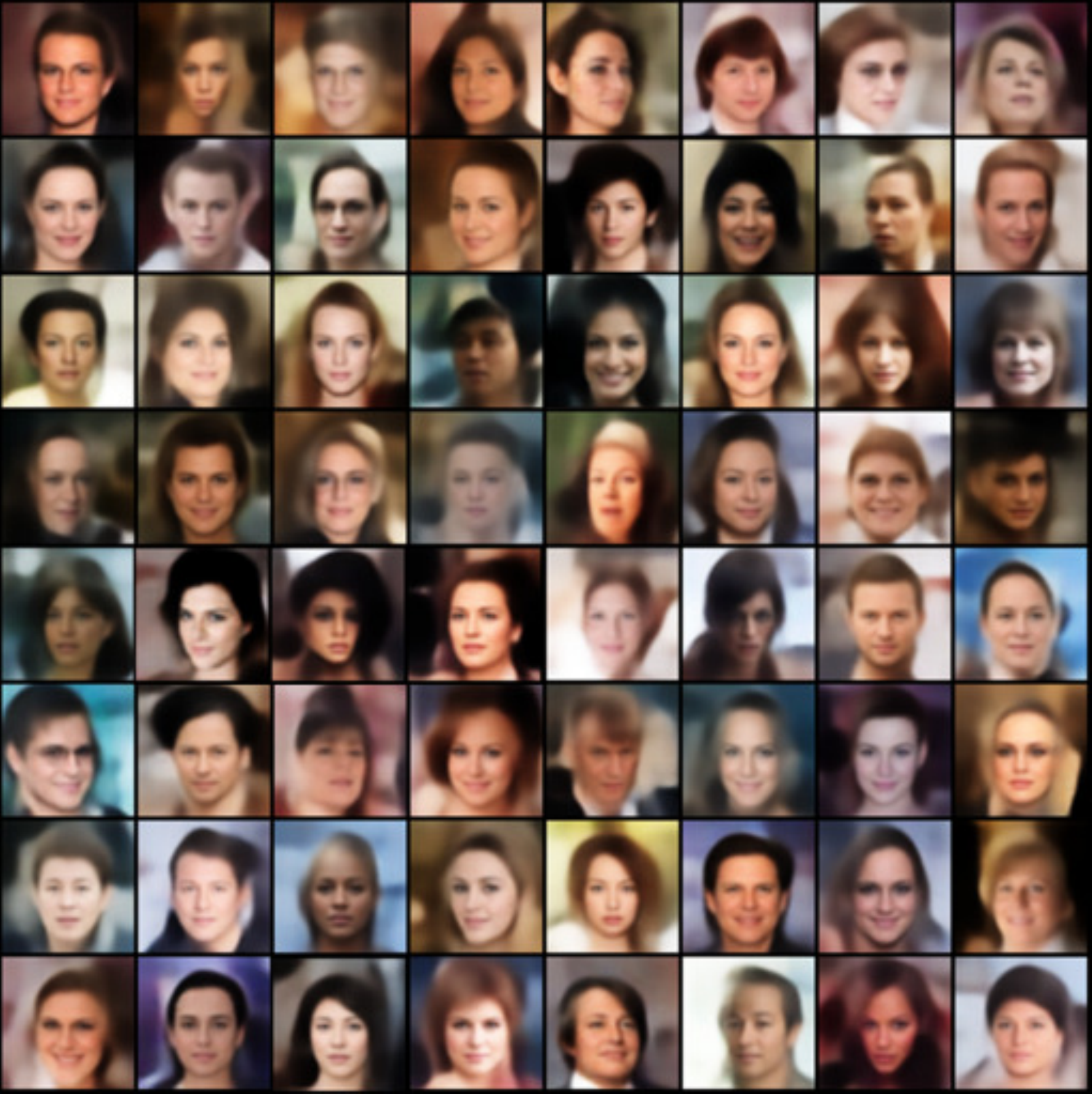}}\vspace{-0.5mm}
  \subfigure[Male]{\includegraphics[scale=0.25]{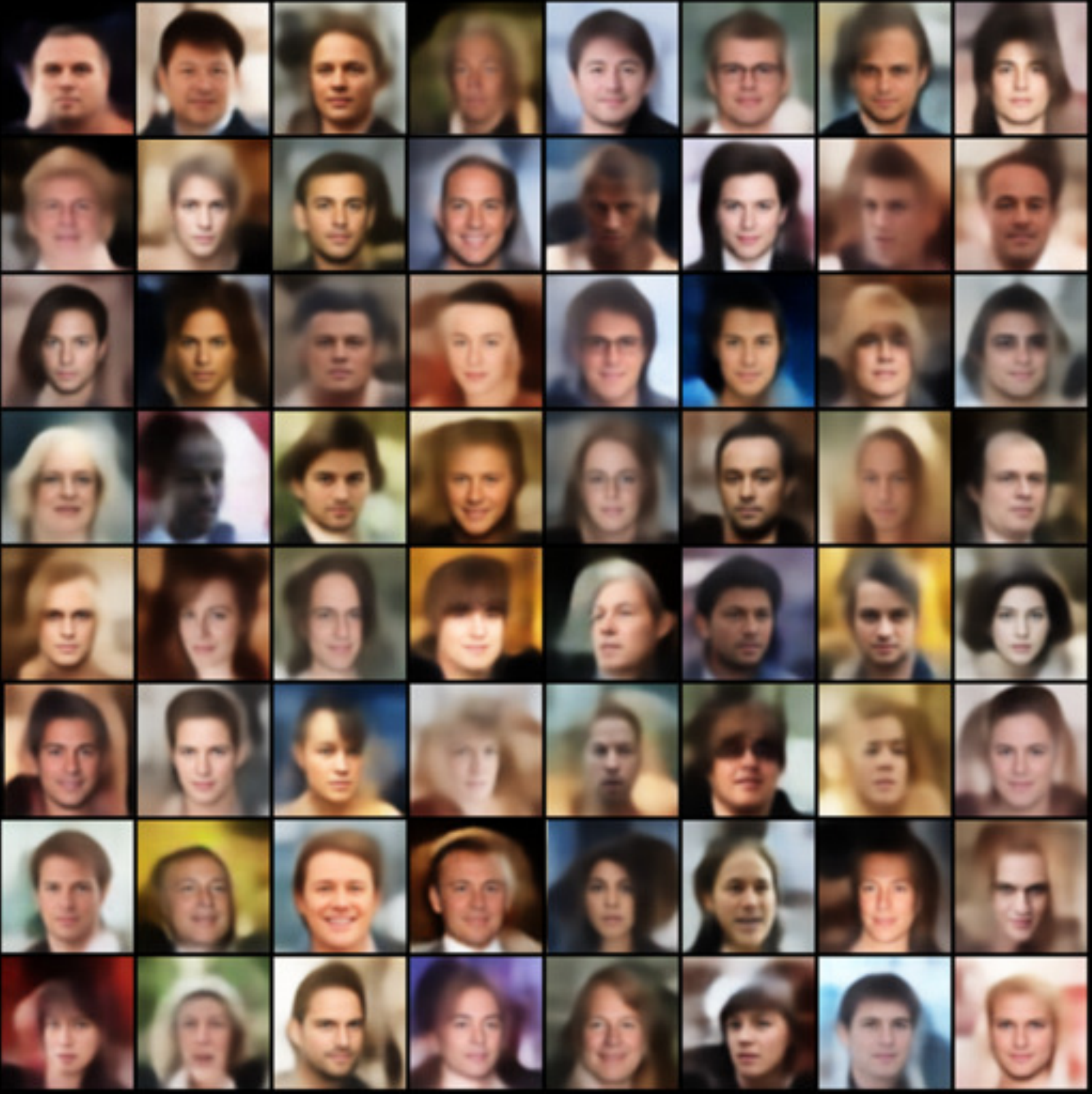}}\vspace{-0.5mm}
  \subfigure[Smiling]{\includegraphics[scale=0.25]{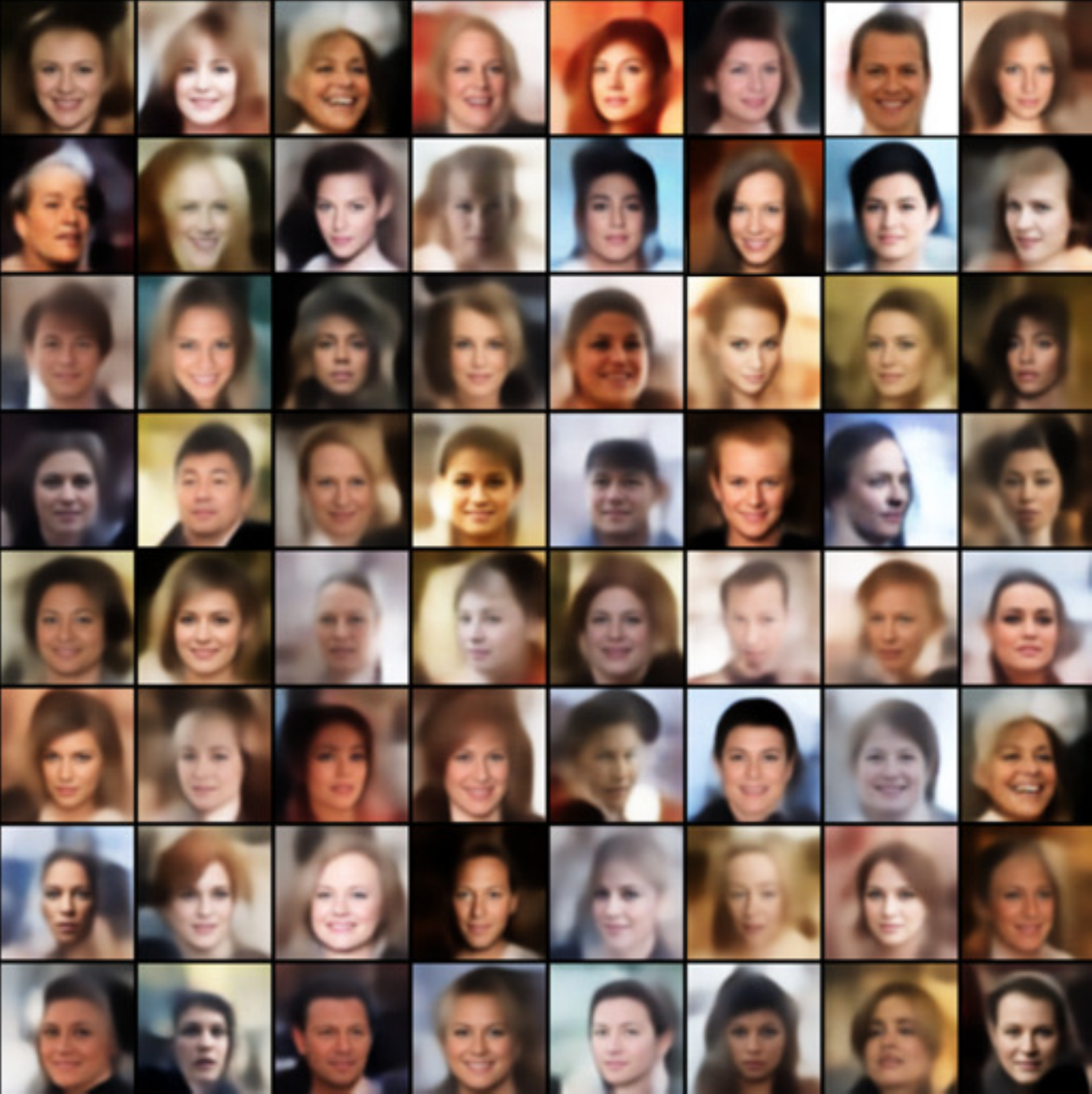}}\vspace{-0.5mm}
  \caption{Generated face images by SMVAE given facial attributes.\label{fig:face}}
\end{figure}

We obtain $1000$ important samples to estimate the probability metrics. Table \ref{mainTable} shows the quatitative results of the proposed SMVAE for each dataset. We can see that the SMVAE outperforms other methods in almost all metrics. The outstanding of SMVAE mainly contributes to the direct modeling of the joint posterior distribution and optimization on a more informative objective. Fig. \ref{fig:mnist}, Fig. \ref{fig:fashion} and Fig. \ref{fig:face} show cross-modality generation of image samples for each domain generated by the SMVAE model. We can see that given the text modality only, the SMVAE can generate corresponding images of good quality. 
\begin{table}[!ht]
\centering
\caption{Statistical results \label{mainTable}}
\setlength\tabcolsep{4pt}
\begin{tabular}{c|l|lll|lll}
\hline
\multicolumn{1}{c|}{DataSet}& Method & \multicolumn{3}{c}{sampled from $p(z|x,y)$}  & \multicolumn{3}{c}{sampled from $p(z|x)$} \\\hline
\multicolumn{1}{l|}{}       &        & $\log p(x,y)$   & $\log p(x)$   & $\log p(x|y)$    & $\log p(x,y)$   & $\log p(x)$      & $\log p(x|y)$  \\\hline
\multirow{4}{*}{FASHION}    & SMVAE  & \textbf{-225.10} & \textbf{-230.81} & -232.85       & \textbf{-225.14}    & \textbf{-232.36}   & -232.84     \\
                            & JMVAE  & -232.70           & -232.40          & \textbf{-230.65}     & -232.94        & -232.63          & \textbf{-230.39} \\
                            & MVAE   & -233.01          & -232.54          & -230.69              & -233.01        & -232.53            & -230.69  \\
                            & MMVAE  & -235.37          & -233.16         & -231.32               & -231.25          &-235.18            & -237.64  \\ \hline
\multirow{4}{*}{MNIST}      & SMVAE  & \textbf{-83.41} & \textbf{-89.38} & \textbf{-86.45}        & \textbf{-83.46}    & \textbf{-89.09}   & \textbf{-86.47}  \\
                            & JMVAE  & -91.03          & -90.962           & -88.43               & -90.76         & -90.69             & -88.69   \\
                            & MVAE   & -90.85          & -90.616           & -88.55               & -90.85         & -90.61             & -88.56   \\
                            & MMVAE   & -91.53         & -91.23           & -90.64                & -91.75          & -91.11             & -90.01   \\
                            & PoMoE   & -90.16         & -90.29           & -                     & -91.36          & -91.77             & -   \\\hline
\multirow{3}{*}{CelebA}     & SMVAE  & \textbf{-6000.95} & \textbf{-6087.91} & \textbf{-6085.31} & \textbf{-6025.38} & \textbf{-6080.86}   & \textbf{-6084.94} \\
                            & JMVAE  & -6238.28          & -6234.54          & -6235.33           & -6242.19       & -6237.97            & -6231.46 \\
                            & MVAE   & -6241.62        & -6237.10           & -6235.36           & -6242.03       & -6236.92            & -6234.95  \\
                            & MMVAE  & -6513.43        & -6560.57           & -6516.14           & -6255.45       & -6284.28            & -6290.74  \\\hline      
\end{tabular}\vspace{-0.5mm}
\end{table}

We further visualize the learned latent representation using tSNE \cite{hinton2002stochastic}. As shown in Fig.\ref{fig:ClatentV}, latent space learned by MVAE method can only produce cohesive latent representation when both modalities are presented. When one modality is missing, representations from their method are distributed irrespective of the semantic category of the data. On the other hand, although the MMVAE method achieves cohesive representation for single-modality posterior, their joint representation is less discriminative. Indicating that using only the combination of uni-modal inference networks is insufficient to capture intermodality co-representation. Nonetheless, our SMVAE method can achieve discriminative latent space for both single- and joint-modality inputs thanks to its ability to exploit shared information from different modalities. 

\begin{figure}[!ht]
  \centering  
  \subfigure[Edge and Facial Landmark Detection ]{\includegraphics[scale=0.36]{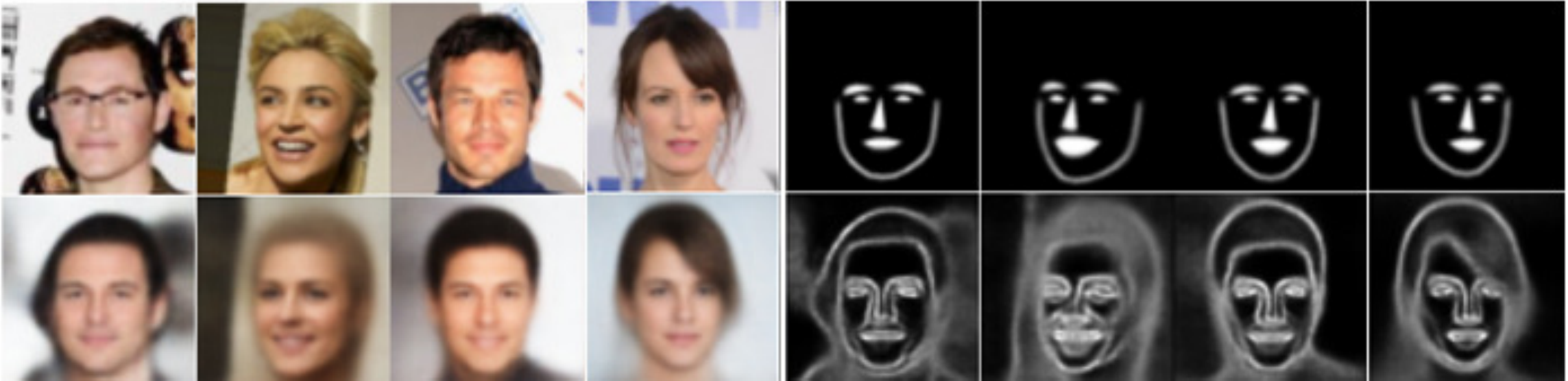}\label{CVStudy-a}}\vspace{-0.5mm}
  \subfigure[Image Recovery from Occlusion]{\includegraphics[scale=0.4]{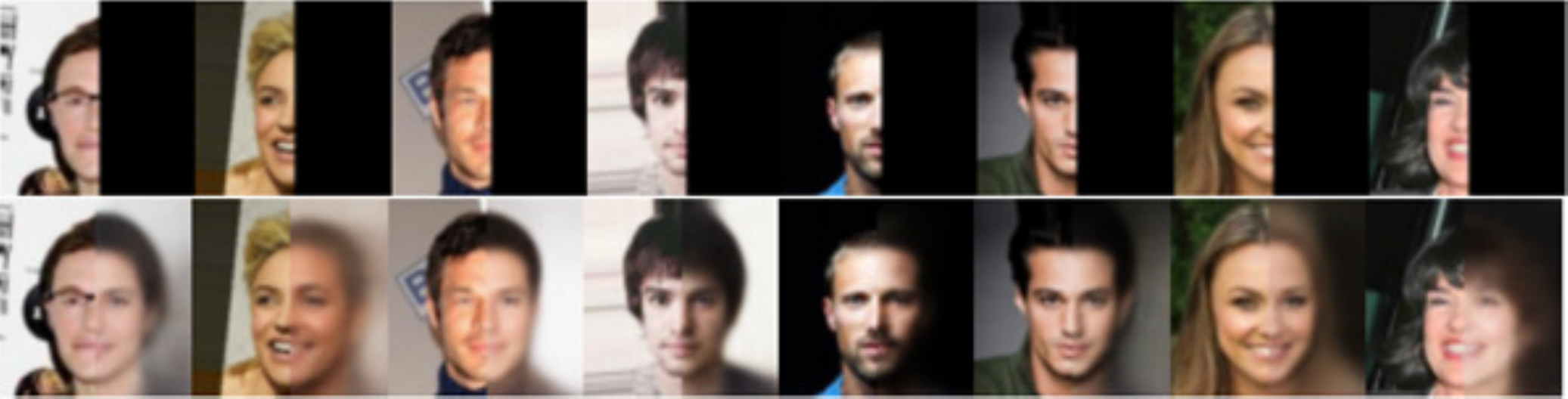}\label{CVStudy-b}}\vspace{-0.5mm}
  \subfigure[Image Painting]{\includegraphics[scale=0.42]{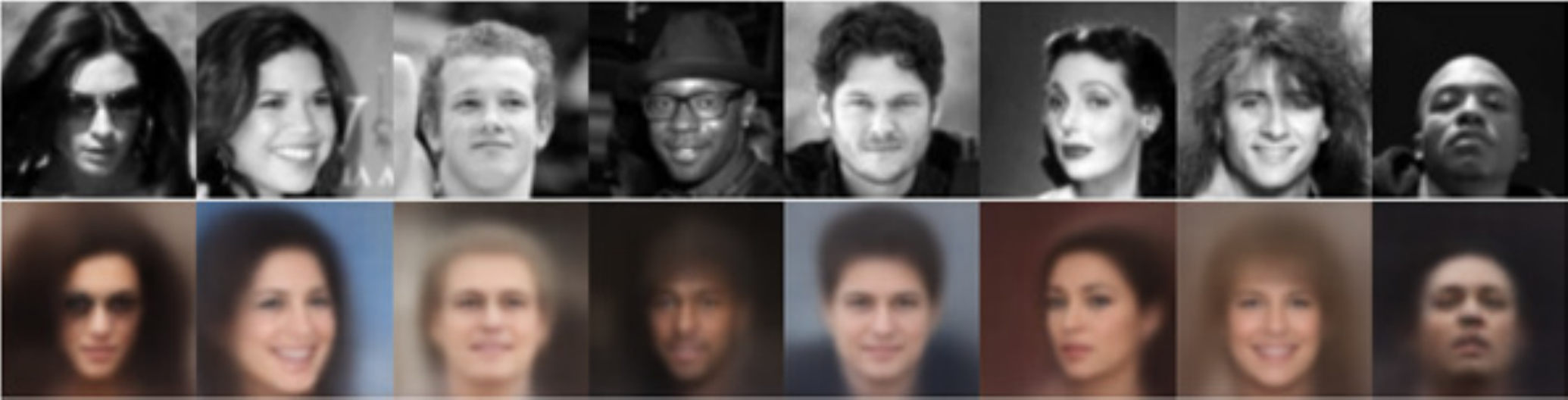}\label{CVStudy-c}}  \vspace{-0.5mm}
  \subfigure[Watermarks Removal]{\includegraphics[scale=0.42]{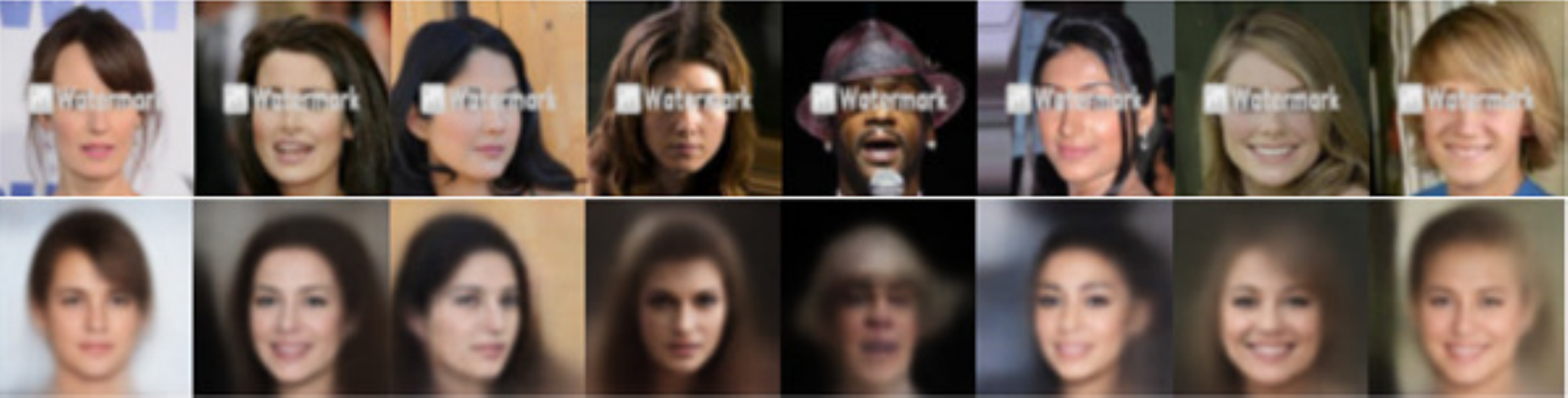}\label{CVStudy-d}}\vspace{-0.5mm}
  \caption{Learning computer vision transformations: (a) Ground truth images are randomly selecetd from CelebA dataset and we present the generated recontructed images, detected edges and facial landmarks. (b) Recontructed color images based on grayscale images. (c) Image recovery from obscured image. (d) Image recovery from watermark images.  \label{vis-CVStudy}}
\end{figure}

\subsection{Case Study: Computer Vision Application}
We demonstrate that our SMVAE is able to learn image transformations including colorization, edge detection, facial landmark segmentation, image completion, and watermark removal. With original image and each transformation as different modalities, we obtain $6$ modalities in total by applying different transformations to the ground-truth images for this multimodal setting. This case study demonstrates the SMVAE's ability to generate in multiple directions and combinations. 

Similar to \cite{wu2018multimodal}, for edge detection, we use Canny detector \cite{canny1986computational} from Scikit-Image module \cite{van2014scikit} to extract edges of the facial image. For facial landmark segmentation, we use Dlib tool \cite{king2009dlib} and OpenCV \cite{bradski2000opencv}. For colorization, we simply convert RGB colors to grayscale. For watermark removal, we add a watermark overlay to the original image. For image completion, we replace half of the image with black pixels. 
Fig.\ref{vis-CVStudy} shows the samples generated from a trained SMVAE model. As can be seen in Fig.\ref{CVStudy-a}, the SMVAE generates a good reconstruction of the facial landmark segmentation and extracted edges. In Fig.\ref{CVStudy-b}, we can see that the SMVAE is able to put reasonable facial color to the input grayscale image. Fig.\ref{CVStudy-c} demonstrates that the SMVAE can recover the image from the watermark and complete the image quite well. The reconstructed right half of the image is basically agreed on the left half of the original image. In Fig.\ref{CVStudy-d}, all traces of the watermark is also removed. Although our reconstructed images suffer from the same blurriness problem that is shared in VAE methods \cite{zhao2017towards}, the SMVAE is able to perform cross-modality generation thanks to its ability to capture share information among modalities.

\begin{figure*}[!ht]
        \centering
        \subfigure[Non-image input]{\includegraphics[scale=0.1]{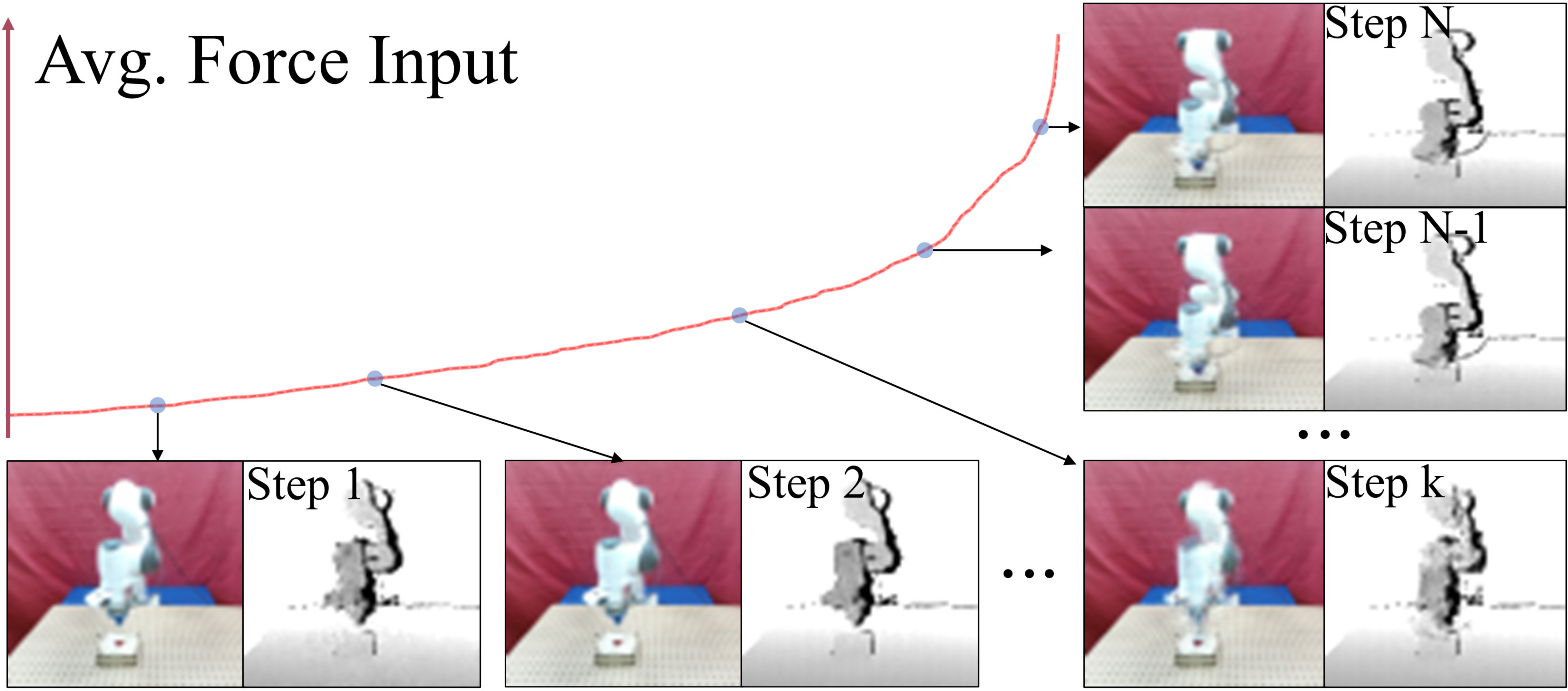} \label{robotics:non-image}} 
        \subfigure[Full information]{\includegraphics[scale=0.4]{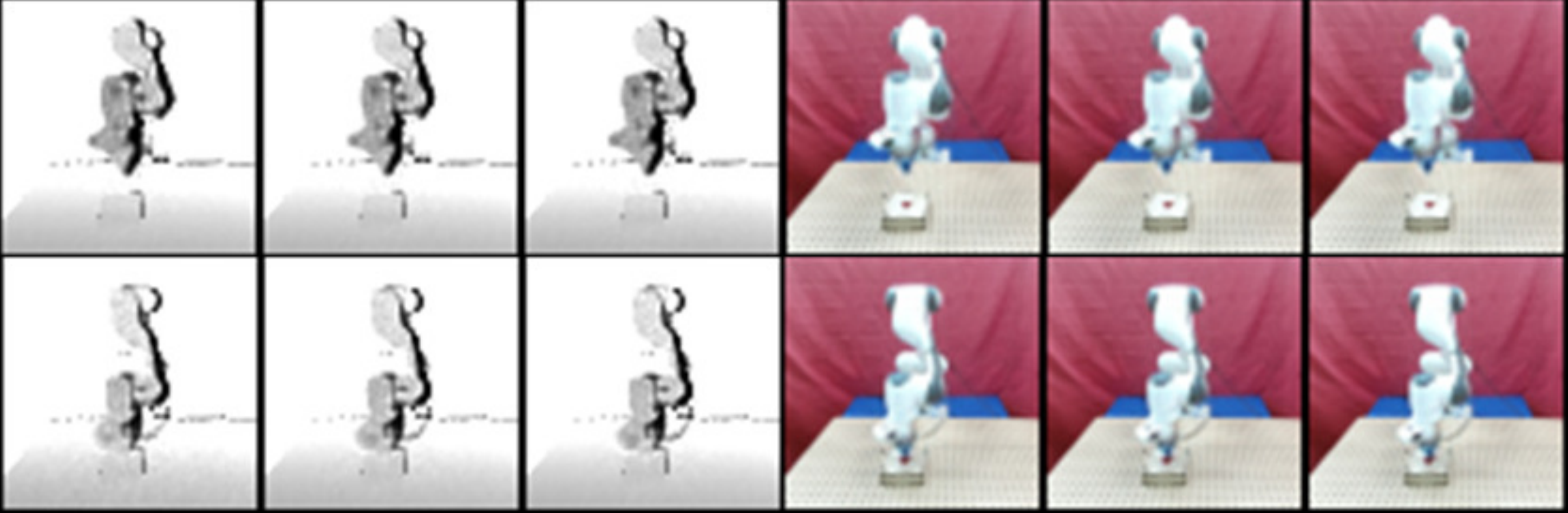}\label{robotics:full}}
        \caption{Learning to reconstruct visual event in robotics scenario. (a) image sampled from SMVAE conditioned on force and action inputs. SMVAE is able to reconstruct different visual modalities that depicts the relative position of the robotic arms to the target box properly. (b) From top to bottom, sampled reconstructed images from a action sequence}
        \label{Vis-Robotics}        
\end{figure*}

\subsection{Case Study: Robotics Control Application}
The second case study shows that our method is readily applicable in robotics control scenarios using Vision\&Touch dataset\cite{liang2021multibench}. We use the SMVAE to learn cross-modality generation from continuous sensory input to images. Emerging human-in-the-loop shared autonomy systems are often equipped with multiple sensors, which pose a high requirement to the model's ability to learn co-representation\cite{lee2020making,luo2021vitac,chen2021multi,selvaggio2021autonomy, newman2022harmonic, li2021toward}. The Vision\&Touch dataset is a real-world robot manipulation dataset that contains visual, tactile, control action, and robot proprioception data which pocess more diverse modalities. The robotic arm attempts to insert the peg located on its tip into the target object. We use a total of $4$ modalities including the depth images, RGB images, the $6$-axis force sensor feedbacks, and the control action given to the robotics arm in each time step. Fig. \ref{robotics:non-image} illustrates that as the robotic arm is not receiving force signals in early steps, reconstruction results of the RGB image show clearly that the arm has no contact with the taget box below. Only when the robotic arm is receiving high force readings, the generated image depicts the contact between the robotics arm and the target box. The quality of the reconstructed rgb and depth images is also differ between partial observation and full observation. While only limited information is observed (i.e., force and action inputs), our method is only able to reconstruct rgb and depth images that can properly reflex the relative posistion between the robotic arm and the target object (Fig. \ref{robotics:non-image}). But when more information is presented, the latent variables can have more comprehensive information about the event and better reconstruction result as we removed the conditional independence assumption (Fig. \ref{robotics:full}). 

\section{Conclusion}\label{Conclusion}
This paper proposes a multimodal generative model by incorporating the set representation learning in the VAE framework.  Unlike the previous multimodal VAE methods, the proposed SMVAE method provides a scalable solution for multimodal data of variable size and permutations. Critically, our model learns the joint posterior distribution directly without additional assumptions for factorization, yielding a more informative objective and the ability to achieve co-representation between modalities. Statistical and visualization results demonstrate that our method excels with other state-of-the-art multimodal VAE methods. Which has high potential in emerging multimodal tasks that need to learn co-representation of diverse data sources while taking missing modality problems or set-input processing problems into consideration. Application on cross-modality reconstruction in robotic dataset further indicates the proposed SMVAE has high potential in emerging multimodal tasks. In the future, we will explore methods that extend the current SMVAE framework to more diverse modalities as well as dynamic multimodal sequences to provide solutions for real-world multimodal applications.     




\bibliography{iclr2023}

\begin{thebibliography}{57}
\providecommand{\natexlab}[1]{#1}
\providecommand{\url}[1]{\texttt{#1}}
\expandafter\ifx\csname urlstyle\endcsname\relax
  \providecommand{\doi}[1]{doi: #1}\else
  \providecommand{\doi}{doi: \begingroup \urlstyle{rm}\Url}\fi

\bibitem[Asano et~al.(2020)Asano, Patrick, Rupprecht, and
  Vedaldi]{asano2020labelling}
Yuki Asano, Mandela Patrick, Christian Rupprecht, and Andrea Vedaldi.
\newblock Labelling unlabelled videos from scratch with multi-modal
  self-supervision.
\newblock \emph{Advances in Neural Information Processing Systems},
  33:\penalty0 4660--4671, 2020.

\bibitem[Atrey et~al.(2010)Atrey, Hossain, El~Saddik, and
  Kankanhalli]{atrey2010multimodal}
Pradeep~K Atrey, M~Anwar Hossain, Abdulmotaleb El~Saddik, and Mohan~S
  Kankanhalli.
\newblock Multimodal fusion for multimedia analysis: a survey.
\newblock \emph{Multimedia systems}, 16\penalty0 (6):\penalty0 345--379, 2010.

\bibitem[Baltru{\v{s}}aitis et~al.(2018)Baltru{\v{s}}aitis, Ahuja, and
  Morency]{baltruvsaitis2018challenges}
Tadas Baltru{\v{s}}aitis, Chaitanya Ahuja, and Louis-Philippe Morency.
\newblock Challenges and applications in multimodal machine learning.
\newblock \emph{The Handbook of Multimodal-Multisensor Interfaces: Signal
  Processing, Architectures, and Detection of Emotion and Cognition-Volume 2},
  pp.\  17--48, 2018.

\bibitem[Beyer et~al.(2020)Beyer, H{\'e}naff, Kolesnikov, Zhai, and
  Oord]{beyer2020we}
Lucas Beyer, Olivier~J H{\'e}naff, Alexander Kolesnikov, Xiaohua Zhai, and
  A{\"a}ron van~den Oord.
\newblock Are we done with imagenet?
\newblock \emph{arXiv preprint arXiv:2006.07159}, 2020.

\bibitem[Bloem-Reddy \& Teh(2020)Bloem-Reddy and Teh]{bloem2020probabilistic}
Benjamin Bloem-Reddy and Yee~Whye Teh.
\newblock Probabilistic symmetries and invariant neural networks.
\newblock \emph{J. Mach. Learn. Res.}, 21:\penalty0 90--1, 2020.

\bibitem[Bowman et~al.(2015)Bowman, Vilnis, Vinyals, Dai, Jozefowicz, and
  Bengio]{bowman2015generating}
Samuel~R Bowman, Luke Vilnis, Oriol Vinyals, Andrew~M Dai, Rafal Jozefowicz,
  and Samy Bengio.
\newblock Generating sentences from a continuous space.
\newblock \emph{arXiv preprint arXiv:1511.06349}, 2015.

\bibitem[Bradski \& Kaehler(2000)Bradski and Kaehler]{bradski2000opencv}
Gary Bradski and Adrian Kaehler.
\newblock Opencv.
\newblock \emph{Dr. Dobb’s journal of software tools}, 3:\penalty0 120, 2000.

\bibitem[Burda et~al.(2015)Burda, Grosse, and
  Salakhutdinov]{burda2015importance}
Yuri Burda, Roger Grosse, and Ruslan Salakhutdinov.
\newblock Importance weighted autoencoders.
\newblock \emph{arXiv preprint arXiv:1509.00519}, 2015.

\bibitem[Canny(1986)]{canny1986computational}
John Canny.
\newblock A computational approach to edge detection.
\newblock \emph{IEEE Transactions on pattern analysis and machine
  intelligence}, \penalty0 (6):\penalty0 679--698, 1986.

\bibitem[Carbonneau et~al.(2018)Carbonneau, Cheplygina, Granger, and
  Gagnon]{CARBONNEAU2018329}
Marc-André Carbonneau, Veronika Cheplygina, Eric Granger, and Ghyslain Gagnon.
\newblock Multiple instance learning: A survey of problem characteristics and
  applications.
\newblock \emph{Pattern Recognition}, 77:\penalty0 329--353, 2018.
\newblock ISSN 0031-3203.
\newblock \doi{https://doi.org/10.1016/j.patcog.2017.10.009}.
\newblock URL
  \url{https://www.sciencedirect.com/science/article/pii/S0031320317304065}.

\bibitem[Chen et~al.(2021)Chen, Lee, and Soh]{chen2021multi}
Kaiqi Chen, Yong Lee, and Harold Soh.
\newblock Multi-modal mutual information (mummi) training for robust
  self-supervised deep reinforcement learning.
\newblock In \emph{2021 IEEE International Conference on Robotics and
  Automation (ICRA)}, pp.\  4274--4280. IEEE, 2021.

\bibitem[Cremer et~al.(2017)Cremer, Morris, and
  Duvenaud]{cremer2017reinterpreting}
Chris Cremer, Quaid Morris, and David Duvenaud.
\newblock Reinterpreting importance-weighted autoencoders.
\newblock \emph{arXiv preprint arXiv:1704.02916}, 2017.

\bibitem[Edelman et~al.(2022)Edelman, Goel, Kakade, and
  Zhang]{edelman2022inductive}
Benjamin~L Edelman, Surbhi Goel, Sham Kakade, and Cyril Zhang.
\newblock Inductive biases and variable creation in self-attention mechanisms.
\newblock In \emph{International Conference on Machine Learning}, pp.\
  5793--5831. PMLR, 2022.

\bibitem[Feng et~al.(2019)Feng, Kong, Zhang, Xue, and Chen]{feng2019good}
Haozhe Feng, Kezhi Kong, Tianye Zhang, Siyue Xue, and Wei Chen.
\newblock Good semi-supervised vae requires tighter evidence lower bound.
\newblock 2019.

\bibitem[Hinton \& Roweis(2002)Hinton and Roweis]{hinton2002stochastic}
Geoffrey~E Hinton and Sam Roweis.
\newblock Stochastic neighbor embedding.
\newblock \emph{Advances in neural information processing systems}, 15, 2002.

\bibitem[Hochreiter \& Schmidhuber(1997)Hochreiter and
  Schmidhuber]{hochreiter1997long}
Sepp Hochreiter and J{\"u}rgen Schmidhuber.
\newblock Long short-term memory.
\newblock \emph{Neural computation}, 9\penalty0 (8):\penalty0 1735--1780, 1997.

\bibitem[Hofer et~al.(2005)Hofer, Odehnal, Pottmann, Steiner, and
  Wallner]{hofer20053d}
Michael Hofer, Boris Odehnal, Helmut Pottmann, Tibor Steiner, and Johannes
  Wallner.
\newblock 3d shape recognition and reconstruction based on line element
  geometry.
\newblock In \emph{Tenth IEEE International Conference on Computer Vision
  (ICCV'05) Volume 1}, volume~2, pp.\  1532--1538. IEEE, 2005.

\bibitem[Hori et~al.(2017)Hori, Hori, Lee, Zhang, Harsham, Hershey, Marks, and
  Sumi]{hori2017attention}
Chiori Hori, Takaaki Hori, Teng-Yok Lee, Ziming Zhang, Bret Harsham, John~R
  Hershey, Tim~K Marks, and Kazuhiko Sumi.
\newblock Attention-based multimodal fusion for video description.
\newblock In \emph{Proceedings of the IEEE international conference on computer
  vision}, pp.\  4193--4202, 2017.

\bibitem[Ioffe \& Szegedy(2015)Ioffe and Szegedy]{ioffe2015batch}
Sergey Ioffe and Christian Szegedy.
\newblock Batch normalization: Accelerating deep network training by reducing
  internal covariate shift.
\newblock In \emph{International conference on machine learning}, pp.\
  448--456. PMLR, 2015.

\bibitem[Khaleghi et~al.(2013)Khaleghi, Khamis, Karray, and
  Razavi]{khaleghi2013multisensor}
Bahador Khaleghi, Alaa Khamis, Fakhreddine~O Karray, and Saiedeh~N Razavi.
\newblock Multisensor data fusion: A review of the state-of-the-art.
\newblock \emph{Information fusion}, 14\penalty0 (1):\penalty0 28--44, 2013.

\bibitem[King(2009)]{king2009dlib}
Davis~E King.
\newblock Dlib-ml: A machine learning toolkit.
\newblock \emph{The Journal of Machine Learning Research}, 10:\penalty0
  1755--1758, 2009.

\bibitem[Kingma \& Welling(2013)Kingma and Welling]{kingma2013auto}
Diederik~P Kingma and Max Welling.
\newblock Auto-encoding variational bayes.
\newblock \emph{arXiv preprint arXiv:1312.6114}, 2013.

\bibitem[LeCun et~al.(1998)LeCun, Bottou, Bengio, and
  Haffner]{lecun1998gradient}
Yann LeCun, L{\'e}on Bottou, Yoshua Bengio, and Patrick Haffner.
\newblock Gradient-based learning applied to document recognition.
\newblock \emph{Proceedings of the IEEE}, 86\penalty0 (11):\penalty0
  2278--2324, 1998.

\bibitem[Lee et~al.(2019)Lee, Lee, Kim, Kosiorek, Choi, and Teh]{lee2019set}
Juho Lee, Yoonho Lee, Jungtaek Kim, Adam Kosiorek, Seungjin Choi, and Yee~Whye
  Teh.
\newblock Set transformer: A framework for attention-based
  permutation-invariant neural networks.
\newblock In \emph{International conference on machine learning}, pp.\
  3744--3753. PMLR, 2019.

\bibitem[Lee et~al.(2020)Lee, Zhu, Zachares, Tan, Srinivasan, Savarese,
  Fei-Fei, Garg, and Bohg]{lee2020making}
Michelle~A Lee, Yuke Zhu, Peter Zachares, Matthew Tan, Krishnan Srinivasan,
  Silvio Savarese, Li~Fei-Fei, Animesh Garg, and Jeannette Bohg.
\newblock Making sense of vision and touch: Learning multimodal representations
  for contact-rich tasks.
\newblock \emph{IEEE Transactions on Robotics}, 36\penalty0 (3):\penalty0
  582--596, 2020.

\bibitem[Li et~al.(2021)Li, Zheng, Fan, and Wang]{li2021toward}
Shufei Li, Pai Zheng, Junming Fan, and Lihui Wang.
\newblock Toward proactive human--robot collaborative assembly: A multimodal
  transfer-learning-enabled action prediction approach.
\newblock \emph{IEEE Transactions on Industrial Electronics}, 69\penalty0
  (8):\penalty0 8579--8588, 2021.

\bibitem[Li \& Turner(2016)Li and Turner]{li2016renyi}
Yingzhen Li and Richard~E Turner.
\newblock R{\'e}nyi divergence variational inference.
\newblock \emph{Advances in neural information processing systems}, 29, 2016.

\bibitem[Liang et~al.(2021)Liang, Lyu, Fan, Wu, Cheng, Wu, Chen, Wu, Lee, Zhu,
  et~al.]{liang2021multibench}
Paul~Pu Liang, Yiwei Lyu, Xiang Fan, Zetian Wu, Yun Cheng, Jason Wu, Leslie
  Chen, Peter Wu, Michelle~A Lee, Yuke Zhu, et~al.
\newblock Multibench: Multiscale benchmarks for multimodal representation
  learning.
\newblock \emph{arXiv preprint arXiv:2107.07502}, 2021.

\bibitem[Liu et~al.(2015)Liu, Luo, Wang, and Tang]{liu2015faceattributes}
Ziwei Liu, Ping Luo, Xiaogang Wang, and Xiaoou Tang.
\newblock Deep learning face attributes in the wild.
\newblock In \emph{Proceedings of International Conference on Computer Vision
  (ICCV)}, December 2015.

\bibitem[Luo et~al.(2021)Luo, Lepora, Martinez-Hernandez, Bimbo, and
  Liu]{luo2021vitac}
Shan Luo, Nathan~F Lepora, Uriel Martinez-Hernandez, Joao Bimbo, and Huaping
  Liu.
\newblock Vitac: Integrating vision and touch for multimodal and cross-modal
  perception.
\newblock \emph{Frontiers in Robotics and AI}, pp.\  134, 2021.

\bibitem[Ma et~al.(2021)Ma, Ren, Zhao, Tulyakov, Wu, and Peng]{ma2021smil}
Mengmeng Ma, Jian Ren, Long Zhao, Sergey Tulyakov, Cathy Wu, and Xi~Peng.
\newblock Smil: Multimodal learning with severely missing modality.
\newblock In \emph{Proceedings of the AAAI Conference on Artificial
  Intelligence}, volume~35, pp.\  2302--2310, 2021.

\bibitem[Murtagh(1991)]{murtagh1991multilayer}
Fionn Murtagh.
\newblock Multilayer perceptrons for classification and regression.
\newblock \emph{Neurocomputing}, 2\penalty0 (5-6):\penalty0 183--197, 1991.

\bibitem[Nagrani et~al.(2020)Nagrani, Sun, Ross, Sukthankar, Schmid, and
  Zisserman]{nagrani2020speech2action}
Arsha Nagrani, Chen Sun, David Ross, Rahul Sukthankar, Cordelia Schmid, and
  Andrew Zisserman.
\newblock Speech2action: Cross-modal supervision for action recognition.
\newblock In \emph{Proceedings of the IEEE/CVF conference on computer vision
  and pattern recognition}, pp.\  10317--10326, 2020.

\bibitem[Newman et~al.(2022)Newman, Aronson, Srinivasa, Kitani, and
  Admoni]{newman2022harmonic}
Benjamin~A Newman, Reuben~M Aronson, Siddhartha~S Srinivasa, Kris Kitani, and
  Henny Admoni.
\newblock Harmonic: A multimodal dataset of assistive human--robot
  collaboration.
\newblock \emph{The International Journal of Robotics Research}, 41\penalty0
  (1):\penalty0 3--11, 2022.

\bibitem[Pandey \& Dukkipati(2017)Pandey and Dukkipati]{pandey2017variational}
Gaurav Pandey and Ambedkar Dukkipati.
\newblock Variational methods for conditional multimodal deep learning.
\newblock In \emph{2017 International Joint Conference on Neural Networks
  (IJCNN)}, pp.\  308--315. IEEE, 2017.

\bibitem[Rahate et~al.(2022)Rahate, Walambe, Ramanna, and
  Kotecha]{rahate2022multimodal}
Anil Rahate, Rahee Walambe, Sheela Ramanna, and Ketan Kotecha.
\newblock Multimodal co-learning: challenges, applications with datasets,
  recent advances and future directions.
\newblock \emph{Information Fusion}, 81:\penalty0 203--239, 2022.

\bibitem[Ramachandran et~al.(2017)Ramachandran, Zoph, and
  Le]{ramachandran2017searching}
Prajit Ramachandran, Barret Zoph, and Quoc~V Le.
\newblock Searching for activation functions.
\newblock \emph{arXiv preprint arXiv:1710.05941}, 2017.

\bibitem[Selvaggio et~al.(2021)Selvaggio, Cognetti, Nikolaidis, Ivaldi, and
  Siciliano]{selvaggio2021autonomy}
Mario Selvaggio, Marco Cognetti, Stefanos Nikolaidis, Serena Ivaldi, and Bruno
  Siciliano.
\newblock Autonomy in physical human-robot interaction: A brief survey.
\newblock \emph{IEEE Robotics and Automation Letters}, 2021.

\bibitem[Shao et~al.(2021)Shao, Bian, Chen, Wang, Zhang, Ji,
  et~al.]{shao2021transmil}
Zhuchen Shao, Hao Bian, Yang Chen, Yifeng Wang, Jian Zhang, Xiangyang Ji,
  et~al.
\newblock Transmil: Transformer based correlated multiple instance learning for
  whole slide image classification.
\newblock \emph{Advances in Neural Information Processing Systems},
  34:\penalty0 2136--2147, 2021.

\bibitem[Shi et~al.(2019)Shi, Paige, Torr, et~al.]{shi2019variational}
Yuge Shi, Brooks Paige, Philip Torr, et~al.
\newblock Variational mixture-of-experts autoencoders for multi-modal deep
  generative models.
\newblock \emph{Advances in Neural Information Processing Systems}, 32, 2019.

\bibitem[Shvetsova et~al.(2022)Shvetsova, Chen, Rouditchenko, Thomas,
  Kingsbury, Feris, Harwath, Glass, and Kuehne]{shvetsova2022everything}
Nina Shvetsova, Brian Chen, Andrew Rouditchenko, Samuel Thomas, Brian
  Kingsbury, Rogerio~S Feris, David Harwath, James Glass, and Hilde Kuehne.
\newblock Everything at once-multi-modal fusion transformer for video
  retrieval.
\newblock In \emph{Proceedings of the IEEE/CVF Conference on Computer Vision
  and Pattern Recognition}, pp.\  20020--20029, 2022.

\bibitem[S{\o}nderby et~al.(2016)S{\o}nderby, Raiko, Maal{\o}e, S{\o}nderby,
  and Winther]{sonderby2016ladder}
Casper~Kaae S{\o}nderby, Tapani Raiko, Lars Maal{\o}e, S{\o}ren~Kaae
  S{\o}nderby, and Ole Winther.
\newblock Ladder variational autoencoders.
\newblock \emph{Advances in neural information processing systems}, 29, 2016.

\bibitem[Studen{\`y} \& Vejnarov{\'a}(1998)Studen{\`y} and
  Vejnarov{\'a}]{studeny1998multiinformation}
Milan Studen{\`y} and Jirina Vejnarov{\'a}.
\newblock The multiinformation function as a tool for measuring stochastic
  dependence.
\newblock In \emph{Learning in graphical models}, pp.\  261--297. Springer,
  1998.

\bibitem[Su et~al.(2015)Su, Maji, Kalogerakis, and Learned-Miller]{su2015multi}
Hang Su, Subhransu Maji, Evangelos Kalogerakis, and Erik Learned-Miller.
\newblock Multi-view convolutional neural networks for 3d shape recognition.
\newblock In \emph{Proceedings of the IEEE international conference on computer
  vision}, pp.\  945--953, 2015.

\bibitem[Sun et~al.(2017)Sun, Shrivastava, Singh, and Gupta]{sun2017revisiting}
Chen Sun, Abhinav Shrivastava, Saurabh Singh, and Abhinav Gupta.
\newblock Revisiting unreasonable effectiveness of data in deep learning era.
\newblock In \emph{Proceedings of the IEEE international conference on computer
  vision}, pp.\  843--852, 2017.

\bibitem[Sutter et~al.(2021)Sutter, Daunhawer, and Vogt]{sutter2021generalized}
Thomas~M. Sutter, Imant Daunhawer, and Julia~E Vogt.
\newblock Generalized multimodal {ELBO}.
\newblock In \emph{International Conference on Learning Representations}, 2021.
\newblock URL \url{https://openreview.net/forum?id=5Y21V0RDBV}.

\bibitem[Suzuki et~al.(2016)Suzuki, Nakayama, and Matsuo]{suzuki2016joint}
Masahiro Suzuki, Kotaro Nakayama, and Yutaka Matsuo.
\newblock Joint multimodal learning with deep generative models.
\newblock \emph{arXiv preprint arXiv:1611.01891}, 2016.

\bibitem[Van~der Walt et~al.(2014)Van~der Walt, Sch{\"o}nberger,
  Nunez-Iglesias, Boulogne, Warner, Yager, Gouillart, and Yu]{van2014scikit}
Stefan Van~der Walt, Johannes~L Sch{\"o}nberger, Juan Nunez-Iglesias,
  Fran{\c{c}}ois Boulogne, Joshua~D Warner, Neil Yager, Emmanuelle Gouillart,
  and Tony Yu.
\newblock scikit-image: image processing in python.
\newblock \emph{PeerJ}, 2:\penalty0 e453, 2014.

\bibitem[Vedantam et~al.(2017)Vedantam, Fischer, Huang, and
  Murphy]{vedantam2017generative}
Ramakrishna Vedantam, Ian Fischer, Jonathan Huang, and Kevin Murphy.
\newblock Generative models of visually grounded imagination.
\newblock \emph{arXiv preprint arXiv:1705.10762}, 2017.

\bibitem[Watanabe(1960)]{watanabe1960information}
Satosi Watanabe.
\newblock Information theoretical analysis of multivariate correlation.
\newblock \emph{IBM Journal of research and development}, 4\penalty0
  (1):\penalty0 66--82, 1960.

\bibitem[Wu \& Goodman(2018)Wu and Goodman]{wu2018multimodal}
Mike Wu and Noah Goodman.
\newblock Multimodal generative models for scalable weakly-supervised learning.
\newblock \emph{Advances in Neural Information Processing Systems}, 31, 2018.

\bibitem[Wu et~al.(2015)Wu, Song, Khosla, Yu, Zhang, Tang, and Xiao]{wu20153d}
Zhirong Wu, Shuran Song, Aditya Khosla, Fisher Yu, Linguang Zhang, Xiaoou Tang,
  and Jianxiong Xiao.
\newblock 3d shapenets: A deep representation for volumetric shapes.
\newblock In \emph{Proceedings of the IEEE conference on computer vision and
  pattern recognition}, pp.\  1912--1920, 2015.

\bibitem[Xiao et~al.(2017)Xiao, Rasul, and Vollgraf]{xiao2017online}
Han Xiao, Kashif Rasul, and Roland Vollgraf.
\newblock Fashion-mnist: a novel image dataset for benchmarking machine
  learning algorithms, 2017.

\bibitem[Yan et~al.(2016)Yan, Yang, Sohn, and Lee]{yan2016attribute2image}
Xinchen Yan, Jimei Yang, Kihyuk Sohn, and Honglak Lee.
\newblock Attribute2image: Conditional image generation from visual attributes.
\newblock In \emph{European conference on computer vision}, pp.\  776--791.
  Springer, 2016.

\bibitem[Zaheer et~al.(2017)Zaheer, Kottur, Ravanbakhsh, Poczos, Salakhutdinov,
  and Smola]{zaheer2017deep}
Manzil Zaheer, Satwik Kottur, Siamak Ravanbakhsh, Barnabas Poczos, Russ~R
  Salakhutdinov, and Alexander~J Smola.
\newblock Deep sets.
\newblock \emph{Advances in neural information processing systems}, 30, 2017.

\bibitem[Zhang et~al.(2021)Zhang, Sidib{\'e}, Morel, and
  M{\'e}riaudeau]{zhang2021deep}
Yifei Zhang, D{\'e}sir{\'e} Sidib{\'e}, Olivier Morel, and Fabrice
  M{\'e}riaudeau.
\newblock Deep multimodal fusion for semantic image segmentation: A survey.
\newblock \emph{Image and Vision Computing}, 105:\penalty0 104042, 2021.

\bibitem[Zhao et~al.(2017)Zhao, Song, and Ermon]{zhao2017towards}
Shengjia Zhao, Jiaming Song, and Stefano Ermon.
\newblock Towards deeper understanding of variational autoencoding models.
\newblock \emph{arXiv preprint arXiv:1702.08658}, 2017.

\end{thebibliography}
\bibliographystyle{iclr2023_conference}

\clearpage
\appendix

\section{Analysis of Independence Assumption and Factorized Objective}\label{Appendix:AnalysisELBO}
Imposing conditional independence assumption to factorize the joint posterior to a number of single-modality posterior trades off the model's capacity to learn co-representation for flexibility against missing modalities. Here we present the detailed derivation for Subsection \ref{subsec:5}:
\begin{equation}
\begin{split}
\mathcal{L} :=ELBO =&\mathbb{E}_{z \sim q_\phi\left(z \mid x_1, \cdots, x_M\right)}\left[\log p_\theta\left(x_1, \cdots, x_M \mid z\right)+\log \frac{p_\theta(z)}{q_\phi\left(z \mid x_1, \cdots, x_M\right)}\right]\\
&p(x_1,\dots,x_M\mid z) = \prod_{i=1}^Mp(x_i\mid z)
\end{split}
\end{equation}

\begin{equation}
    \label{Extend1}
    \begin{split}
    \mathcal{L}&=\mathbb{E}_{q_\phi(\mathbf{z}\mid\mathbb{X})}\log \frac{p_\theta(\mathbb{X}, \mathbf{z})}{q_\phi(\mathbf{z}\mid\mathbb{X})}\\
    &=\mathbb{E}_{q_\phi(\mathbf{z}\mid\mathbb{X})}\log \frac{p_\theta(\mathbb{X}, \mathbf{z})\cdot p_\theta(z)\prod_{i=1}^Mp_\theta(x_i\mid z)}{q_\phi(\mathbf{z}\mid\mathbb{X})\cdot p_\theta(z)\prod_{i=1}^Mp_\theta(x_i\mid z)} \\
    &=\mathbb{E}_{q_\phi(\mathbf{z}\mid\mathbb{X})} \left[\log \frac{p_\theta(z)\prod_{i=1}^Mp_\theta(x_i\mid z)}{q_\phi(\mathbf{z}\mid\mathbb{X})}+\log \frac{p_\theta(\mathbb{X}, \mathbf{z})}{p_\theta(z)\prod_{i=1}^Mp(x_i\mid z)}\right] \\
    &=\mathcal{L}_{CI} +\mathbb{E}_{q_\phi(\mathbf{z}\mid\mathbb{X})} \left[\log \frac{p_\theta(\mathbb{X}\mid \mathbf{z})}{\prod_{i=1}^Mp_\theta(\mathbf{x}_i\mid z)}\right]
    \end{split}
\end{equation}
\begin{equation}
    \label{Extend2}
    \begin{split}
    \mathbb{E}_{q(\mathbb{X})}\left[\mathcal{L}\right]&= \mathbb{E}_{q(\mathbb{X})}\left[\mathcal{L}_{CI}\right]+ \mathbb{E}_{q(\mathbb{X})}\mathbb{E}_{q_\phi(\mathbf{z}\mid\mathbb{X})} \left[ \log \frac{p_\theta(\mathbb{X}\mid \mathbf{z})}{\prod_{i=1}^Mp_\theta(x_i\mid \mathbf{z})}\right] \\
    &= \mathbb{E}_{q(\mathbb{X})}\left[\mathcal{L}_{CI}\right]+ \iint{q(\mathbb{X})}{q_\phi(\mathbf{z}\mid\mathbb{X})} \left[\log \frac{p_\theta(\mathbb{X}\mid \mathbf{z})}{\prod_{i=1}^Mp_\theta(x_i\mid \mathbf{z})}\right]\mathrm{d}z\mathrm{d}\mathbb{X} \\
    &= \mathbb{E}_{q(\mathbb{X})}\left[\mathcal{L}_{CI}\right]+ \iint \frac{q(\mathbb{X})q_\phi(\mathbf{z}\mid\mathbb{X})}{p_\theta(\mathbb{X}\mid \mathbf{z})}p_\theta(\mathbb{X}\mid \mathbf{z}) \left[\log \frac{p_\theta(\mathbb{X}\mid \mathbf{z})}{\prod_{i=1}^Mp_\theta(x_i\mid \mathbf{z})}\right]\mathrm{d}z\mathrm{d}\mathbb{X} \\
    &= \mathbb{E}_{q(\mathbb{X})}\left[\mathcal{L}_{CI}\right]+ \int \frac{q(\mathbb{X})q_\phi(\mathbf{z}\mid\mathbb{X})}{p_\theta(\mathbb{X}\mid \mathbf{z})}\int p_\theta(\mathbb{X}\mid \mathbf{z}) \left[\log \frac{p_\theta(\mathbb{X}\mid \mathbf{z})}{\prod_{i=1}^Mp_\theta(x_i\mid \mathbf{z})}\right]\mathrm{d}\mathbb{X}\mathrm{d}z \\
    &= \mathbb{E}_{q(\mathbb{X})}\left[\mathcal{L}_{CI}\right]+ \mathbb{E}_{z\sim\frac{q(\mathbb{X}) q_\phi(z\mid\mathbb{X})}{p_\theta(\mathbb{X}\mid z)}}\left[ \underset{\text{conditional total correlation}}{\underbrace{\mathbb{E}_{\mathbb{X}\sim p_\theta(\mathbb{X}\mid \mathbf{z})}\left[\log \frac{p_\theta(\mathbb{X}\mid \mathbf{z})}{\prod_{i=1}^Mp_\theta(x_i\mid \mathbf{z})}\right]}}\right]
    \end{split}
\end{equation}
, where $\mathbb{X} \equiv (x_1,\cdots,x_M)$. In another viewpoint, literatures have pointed out that a larger gap between the objective and the actual log likelihood of the input data will hinders the performance of VAE. Several works thus proposed a tighter ELBO to facilitate the learning process \cite{feng2019good, burda2015importance, cremer2017reinterpreting,li2016renyi}. 

Empirical results from \cite{suzuki2016joint} also show that without further constraints, the inferred latent variables from incomplete modalities will be impaired and causes degradation to to the generative samples's quality. Which indicates that $z$ is not completely independent to $x_j$ given only $x_i$. So it is fair to say that $p(z|x_1, \cdots, x_N) \ne p(z|x_i)$. This insight motivate us to drop the conditional independence assumption and find another way to achieve a scalable multimodal model. Different from factorization method, our SMVAE optimize on a tighter lower bound of the data's log-likelihood while being able to exploit corelations between different modalities.

\section{Detailed Training Configurations and Architectures of Encoder and Decoder} \label{Appendix-Arch}
For MNIST and FASHION dataset, we consider images as $\mathbf{x}\in\mathbb{R}^{28\times{28}}$ and the corresponding labels as one-hot vectors $\mathbf{y}\in\{0,1\}^{10}$. The detailed architectures of encoders and decoders for these datasets are listed in Table \ref{mnist-ed}, and Table \ref{fashion-ed}. The size of latent variables is set to $64$, i.e., $L=64$, the weighting coefficient for images is $1.0$ while that of the label is set to $10.0$, total number of epochs is set to $200$, and mini-batch size is set to $100$ for all methods. The learning rate is set to $5e^{-4}$ for our SMVAE method while in other comparing methods, it is set to $1e^{-3}$ according to setting reported in their papers. We let $\beta$ to anneal for $100$ out of $200$ epochs. 

For CelebA dataset, we use crop central face images and resize the image to $64\times{64}$. The resized images are regarded as $\mathbf{x}\in\mathbb{R}^{3\times{64}\times{64}}$ while the corresponding attributes are also processed as one-hot vectors $\mathbf{y}\in\{0,1\}^{18}$. The architectures of encoders and decoders are listed in Table \ref{celeba-ed}. The size of latent variables is set to $100$, the weighting coefficient for images is $1.0$ and that of the label is set to $10.0$, total number of epochs is set to $200$, and mini-batch size is set to $100$ for all methods. The learning rate is set to $5e^{-4}$ for our SMVAE method while in other comparing methods, it is set to $1e^{-3}$ according to setting reported in their papers. We let $\beta$ to anneal for $100$ out of $200$ epochs.
\begin{table}[h!]
\renewcommand\arraystretch{1.2}
\setlength{\tabcolsep}{7mm}
\centering
\begin{threeparttable}
\begin{tabular}{l|l|l}
\hline
Input $K\in\mathbb{R}^{m\times{512}}$& Input $V\in\mathbb{R}^{m\times{512}}$ & Input $Q\in\mathbb{R}^{1\times{512}}$\\
FC.$512$.Swish  & FC.$512$.Swish  & FC.$512$.Swish\\\hline
\multicolumn{3}{c}{Split($512//h$)}         \\
\multicolumn{3}{c}{Softmax$(QK^T/\sqrt{V})\times{V}$}\\
\multicolumn{3}{c}{Skip-Connection, LayerNorm}\\
\multicolumn{3}{c}{FC.L$\times$2}\\\hline
\end{tabular}
\begin{tablenotes}
\item [1] FC.$N$ denotes a fully connected layer with $N$ hidden units and outputs a $N$ dimensional vector.
\item [2] h denotes number of attention heads.
\item [3] $L$ denotes the size of latent space
\end{tablenotes} 
\end{threeparttable}
\caption{Set Encoder architecture. \label{SE}}
\end{table}

\begin{table}[h!]
\renewcommand\arraystretch{1.2}
\centering
\begin{threeparttable}
\begin{tabular}{c|c|c|c}
\hline
Encoder-Image & Decoder-Image  & Encoder-Label & Decoder-Label\\  \hline
Input $\mathbf{x}\in\mathbb{R}^{1\times{28}\times{28}}$&Input $\mathbf{z}\in\mathbb{R}^{L}$&Input $\mathbf{y}\in\mathbb{R}^{1\times{10}}$&Input $\mathbf{z}\in\mathbb{R}^{L}$\\
FC.$512$.Swish& FC.$512$.Swish        & FC.$512$.Swish & FC.$512$.Swish\\
   -          & FC.$1\times{28}\times{28}$&   -       & FC.$10$\\\hline
\end{tabular}
\begin{tablenotes}
\item [1] FC.$N$ denotes a fully connected layer with $N$ hidden units and outputs a $N$ dimensional vector.
\item [2] Swish denotes swish activation function \cite{ramachandran2017searching}.
\item [3] $L$ denotes the size of latent space
\end{tablenotes} 
\caption{Encoder and decoder architecture for MNIST dataset. \label{mnist-ed}}
\end{threeparttable}
\end{table}

\begin{table}[h!]
\renewcommand\arraystretch{1.2}
\begin{threeparttable}
\centering
\begin{tabular}{c|c|c|c}
\hline
Encoder-Image         & Decoder-Image         & Encoder-Label & Decoder-Label            \\\hline
Input $\mathbf{x}\in\mathbb{R}^{1\times{28}\times{28}}$&Input $\mathbf{z}\in\mathbb{R}^{L}$&Input $\mathbf{y}\in\mathbb{R}^{1\times{10}}$&Input $\mathbf{z}\in\mathbb{R}^{L}$\\
Conv.$64$.Swish       & FC.$512$.Swish       & FC.$512$.Swish & FC.$512$.Swish           \\
Conv.$128$.Swish      & FC.$128\times 7\times 7$.Swish &      & FC.$10$                  \\
FC.$512$.Swish        & DeConv.$64$.Swish & - &-\\
      -               & DeConv.$1$       & -&-\\\hline
\end{tabular}
\begin{tablenotes}
\item [1] Conv.$N$ denotes a convolutional layer with $N$ kernels and $N$ output channels.
\item [2] DeConv.$N$ denotes a deconvolutional layer with $N$ kernels and $N$ output channels.
\item [3] BN denotes a batch normalization layer \cite{ioffe2015batch}.
\item [4] Swish denotes swish activation function \cite{ramachandran2017searching}.
\item [5] All the convolutional and deconvolutional layers has strides set to $4$ and padding set to $2$. 
\item [6] Other abbreviations can refer to Table \ref{mnist-ed}.
\end{tablenotes} 
\caption{Encoder and decoder architecture for experiments on Fashion dataset.\label{fashion-ed}}
\end{threeparttable}
\end{table}

\begin{table}[h!]
\renewcommand\arraystretch{1.2}
\centering
\begin{threeparttable}
\begin{tabular}{c|c|c|c}
\hline
Encoder-Image       & Decoder-Image            &Encoder-Attributes &Decoder-Attributes\\\hline
Input $\mathbf{x}\in\mathbb{R}^{3\times{64}\times{64}}$&Input $\mathbf{z}\in\mathbb{R}^{L}$&Input $\mathbf{y}\in\mathbb{R}^{1\times{18}}$&Input $\mathbf{z}\in\mathbb{R}^{L}$\\
Conv.$32$         & FC.$256\times{5}\times{5}$.Swish&FC.$512$.BN.Swish&FC.$512$.BN.Swish\\
Conv.$64$.BN.Swish  & DeConv.$128$.Swish       &FC.$512$.BN.Swish &FC.$512$.BN.Swish\\
Conv.$128$.BN.Swish & DeConv.$64$.Swish        &FC.$512$.BN       &FC.$512$.BN.Swish\\
Conv.$256$.BN.Swish & DeConv.$32$.Swish        &  -                &FC.$18$.BN.Swish \\
FC.$512$.Swish      & DeConv.$3$.Swish         &  -  &  -    \\\hline
\end{tabular}
\begin{tablenotes}
\item [1] abbreviations can refer to Table \ref{mnist-ed} and Table \ref{fashion-ed} 
\end{tablenotes} 
\caption{Encoder and decoder architecture for experiments on CelebA dataset. \label{celeba-ed}}
\end{threeparttable}
\end{table}

For the computer vision application setting, we use a SMVAE model with $250$ latent size. The learning rate is set to $10^{-4}$. Regular image, obscured image and image with watermarks are is regarded as image of $\mathbf{x}\in\mathbb{R}^{3\times{64}\times{64}}$ while the other image modalities are regarded as $\mathbf{x}\in\mathbb{R}^{1\times{64}\times{64}}$. Mini-batch size is set to $100$, and learning rate is set to $5e^{-5}$. We let $\beta$ to anneal for $50$ epochs out of $100$ epochs.  Table \ref{vit-ed} shows the detailed architectures for encoder and decoder for each modality. 

\begin{table}[h!]
\renewcommand\arraystretch{1.2}
\centering
\begin{threeparttable}
\begin{tabular}{c|c|c|c}
\hline
Encoder-Img/Obs/Wm &Decoder-Img/Obs/Wm &Encode-Gray/Edg/Mask &Decoder-Gray/Edge/Mask\\\hline
Input $\mathbf{x}\in\mathbb{R}^{3\times{64}\times{64}}$&Input $\mathbf{z}\in\mathbb{R}^{L}$&Input $\mathbf{x}\in\mathbb{R}^{3\times{64}\times{64}}$&Input $\mathbf{z}\in\mathbb{R}^{L}$\\
Conv.$32$           & FC.$256\times{5}\times{5}$.Swish &Conv.$32$           & FC.$256\times{5}\times{5}$.Swish\\
Conv.$64$.BN.Swish  & DeConv.$128$.Swish               &Conv.$64$.BN.Swish  & DeConv.$128$.Swish  \\
Conv.$128$.BN.Swish & DeConv.$64$.Swish                &Conv.$128$.BN.Swish & DeConv.$64$.Swish \\
Conv.$256$.BN.Swish & DeConv.$32$.Swish                &Conv.$256$.BN.Swish & DeConv.$32$.Swish   \\
FC.$512$.Swish      & DeConv.$3$.Swish                 & FC.$512$.Swish      & DeConv.$3$.Swish \\\hline
\end{tabular}
\begin{tablenotes}
\item [1] Obs denotes obscured image, Img denotes regular RGB image, Wm denotes image with watermarks.
\item [2] abbreviations can refer to Table \ref{mnist-ed} and Table \ref{fashion-ed} .
\end{tablenotes} 
\caption{Encoder and decoder architecture for experiments on computer vision transformations. \label{vit-ed}}
\end{threeparttable}
\end{table}

For the robotics application setting, there are a total of $150$ trajectories in the Vision\&Touch dataset, we use $70\%$ of samples independently drawn from the dataset as the train set and the remaining $30\%$ as the test set. The rgb images are regarded as $\mathbf{x}_1\in\mathbb{R}^{3\times{64}\times{64}}$, the depth images are regarded as $\mathbf{x}_2\in\mathbb{R}^{1\times{64}\times{64}}$, action commands are regarded as $\mathbf{x}_3\in\mathbb{R}^{4}$, and force sensor inputs are regarded as $\mathbf{x}_4\in\mathbb{R}^{6\times{32}}$. We train a SMVAE model with $64$ latent size, and set the learning rate to $1e^{-3}$. We allow $\beta$ to anneal for $150$ epochs out of $200$ epochs. The architecture for the encoder and decoder are listed in Table \ref{robotics-encoders} and Table \ref{robotics-decoders}

\begin{table}[h!]
\renewcommand\arraystretch{1.2}
\centering
\begin{threeparttable}
\begin{tabular}{c|c|c|c}
\hline
Encoder-Img/Obs/Wm &Decoder-Img/Obs/Wm &Encoder-Gray/Edg/Mask &Decoder-Gray/Edge/Mask\\\hline
Input $\mathbf{x}\in\mathbb{R}^{3\times{64}\times{64}}$&Input $\mathbf{z}\in\mathbb{R}^{L}$&Input $\mathbf{x}\in\mathbb{R}^{3\times{64}\times{64}}$&Input $\mathbf{z}\in\mathbb{R}^{L}$\\
Conv.$32$           & FC.$256\times{5}\times{5}$.Swish &Conv.$32$           & FC.$256\times{5}\times{5}$.Swish\\
Conv.$64$.BN.Swish  & DeConv.$128$.Swish               &Conv.$64$.BN.Swish  & DeConv.$128$.Swish  \\
Conv.$128$.BN.Swish & DeConv.$64$.Swish                &Conv.$128$.BN.Swish & DeConv.$64$.Swish \\
Conv.$256$.BN.Swish & DeConv.$32$.Swish                &Conv.$256$.BN.Swish & DeConv.$32$.Swish   \\
FC.$512$.Swish      & DeConv.$3$.Swish                 & FC.$512$.Swish      & DeConv.$3$.Swish \\\hline
\end{tabular}
\begin{tablenotes}
\item [1] Obs denotes obscured image, Img denotes regular RGB image, Wm denotes image with watermarks.
\item [2] Abbreviations can refer to Table \ref{mnist-ed} and Table \ref{fashion-ed}.
\end{tablenotes} 
\caption{Encoder and decoder architecture for experiments on computer vision transformations. \label{cv-ed}}
\end{threeparttable}
\end{table}

\begin{table}[h!]
\renewcommand\arraystretch{1.2}
\centering
\begin{threeparttable}
\begin{tabular}{c|c|c|c}
\hline
Encoder-Image       &Encoder-Depth             &Encoder-Action          &Encoder-Force  \\\hline
Input $\mathbf{x}\in\mathbb{R}^{3\times{64}\times{64}}$&Input $\mathbf{x}\in\mathbb{R}^{1\times{64}\times{64}}$&Input $\mathbf{x}_3\in\mathbb{R}^{4}$ &Input $\mathbf{x}_4\in\mathbb{R}^{6\times{32}}$\\
Conv.$32$           & Conv.$32$                &FC.$512$.Swish          & CausalConv1D.$16$\\
Conv.$64$.BN.Swish  & Conv.$64$.BN.Swish       &FC.$512$                & CausalConv1D.$32$ \\
Conv.$128$.BN.Swish & Conv.$128$.BN.Swish      &-                       & CausalConv1D.$64$ \\
Conv.$256$.BN.Swish & Conv.$256$.BN.Swish      &-                       & CausalConv1D.$128$\\
FC.$512$.Swish      & FC.$512$.Swish           &-                       & FC.$512$ \\\hline
\end{tabular}
\begin{tablenotes}
\item [1] Other abbreviations can refer to Table \ref{mnist-ed} and Table \ref{fashion-ed}.
\end{tablenotes} 
\caption{Encoder architecture for experiments on robotics dataset. \label{robotics-encoders}}
\end{threeparttable}
\end{table}

\begin{table}[h!]
\renewcommand\arraystretch{1.2}
\centering
\begin{threeparttable}
\begin{tabular}{c|c|c|c}
\hline
Decoder-Image       &Decoder-Depth             &Decoder-Action          &Decoder-Force  \\\hline
Input $\mathbf{z}\in\mathbb{R}^{L}$& Input $\mathbf{z}\in\mathbb{R}^{L}$ &Input $\mathbf{z}\in\mathbb{R}^{L}$ &Input $\mathbf{z}\in\mathbb{R}^{L}$\\
FC.$256\times{5}\times{5}$.Swish&FC.$256\times{5}\times{5}$.Swish &FC.$512$.Swish            & FC.$128\times{2}$.Swish\\
DeConv.$128$.Swish  & DeConv.$128$.Swish              &FC.$4$                                & DeConv1D.$64$  \\
DeConv.$64$.Swish   & DeConv.$64$.Swish               &-                                     & DeConv1D.$32$ \\
DeConv.$32$.Swish   & DeConv.$32$.Swish               &-                                     & DeConv1D.$16$ \\
DeConv.$3$.Swish    & DeConv.$3$.Swish                &-                                     & DeConv1D.$6$ \\\hline
\end{tabular}
\begin{tablenotes}
\item [1] Abbreviations can refer to Table \ref{mnist-ed} and Table \ref{fashion-ed}.
\end{tablenotes} 
\caption{Decoder architecture for experiments on robotics dataset. \label{robotics-decoders}}
\end{threeparttable}
\end{table}

\section{Appendix 3. Probabilistic Symmetry} \label{Apppendix:ProbabilisticSymmetry}
Recent work on probabilistic symmetries has established a correspondence between functional symmetries with probabilistic symmetries. Here we quote some contents from \cite{bloem2020probabilistic}. As one of their main results, for an exchangeable input $\mathbf{X}_n$, the conditional distribution of an output $\mathbf{Y}$ is invariant to permutations of $\mathbf{X}_n$ if and only if there exists a function $f$ such that:
\begin{equation}
\label{permutation-invariant-representation}
(\mathbf{X}_n,\mathbf{Y}) \overset{a.s.}{=} (x_set,f(\eta,\mathbb{M}_{\mathbf{X}_n}))
\end{equation}
, where $\mathbb{M}_{\mathbf{X}_n}$ is some aggregation function for $\mathbf{X}_n$ in summation form, $\eta$ is a noise variable that acts as a generic source of stochasticity and $\overset{a.s.}{=}$ denotes equal almost surely. An example of this is the noise-outsourced functional representation which is a general version of the reparameterization trick. The equation holds as long as $\mathbb{M}_{\mathbf{X}_n}$ is a function that removes the order of elements in $\mathbf{X}_n$. We adopt the above process as support for designing the set representation in our probabilistic model. More detailed explanations can refer to its original paper \cite{bloem2020probabilistic}.



\end{document}